\pdfoutput=1

\documentclass[11pt]{article}

\usepackage[final]{acl}

\usepackage{times}
\usepackage{latexsym}

\usepackage[T1]{fontenc}

\usepackage[utf8]{inputenc}

\usepackage{microtype}

\usepackage{inconsolata}

\usepackage{graphicx}

\DeclareUnicodeCharacter{03B2}{$\beta$}
\DeclareUnicodeCharacter{2212}{-}
\usepackage[acronym]{glossaries}
\glsdisablehyper
\newacronym{ours}{R2E}{Retrieve to Explain}
\usepackage{enumitem}
\usepackage{amssymb}
\usepackage{booktabs}
\usepackage{multirow}
\usepackage{array}
\usepackage{algorithm}
\usepackage{algpseudocode}
\let\oldacrshort\acrshort
\renewcommand{\acrshort}[1]{\texorpdfstring{\oldacrshort{#1}}{}}

\title{Retrieve to Explain: Evidence-driven Predictions for Explainable Drug Target Identification}

\author{%
  Ravi Patel \qquad
  Angus Brayne \qquad
  Rogier Hintzen \qquad \\
  \textbf{Daniel Jaroslawicz} \qquad
  \textbf{Georgiana Neculae} \qquad
  \textbf{Dane Corneil} \\
  BenevolentAI, London, United Kindgom \\
  \texttt{\{ravi.patel, dane.corneil\}@benevolent.ai}
}

\begin{document}
\maketitle
\begin{abstract}
Language models hold incredible promise for enabling scientific discovery by synthesizing massive research corpora. Many complex scientific research questions have multiple plausible answers, each supported by evidence of varying strength. However, existing language models lack the capability to quantitatively and faithfully compare answer plausibility in terms of supporting evidence. To address this, we introduce \acrfull{ours}, a retrieval-based model that scores and ranks all possible answers to a research question based on evidence retrieved from a document corpus. The architecture represents each answer only in terms of its supporting evidence, with the answer itself masked. This allows us to extend feature attribution methods such as Shapley values, to transparently attribute answer scores to supporting evidence at inference time. The architecture also allows incorporation of new evidence without retraining, including non-textual data modalities templated into natural language. We developed \acrshort{ours} for the challenging scientific discovery task of drug target identification, a human-in-the-loop process where failures are extremely costly and explainability paramount. When predicting whether drug targets will subsequently be confirmed as efficacious in clinical trials, \acrshort{ours} not only matches non-explainable literature-based models but also surpasses a genetics-based target identification approach used throughout the pharmaceutical industry.
\end{abstract}

\section{Introduction}
\label{intro}

Language models can act as knowledge bases, supplying answers to factual user queries using only the learned parameters \cite{petroni2019language,brayne-etal-2022-masked}. They can also be provided with access to searchable knowledge bases for retrieval-augmented question answering \cite{chen2017reading,lewis2020retrieval,izacard2021leveraging}.

\begin{figure}[t!]
\setlength{\belowcaptionskip}{-8pt}
\includegraphics[width=\linewidth]{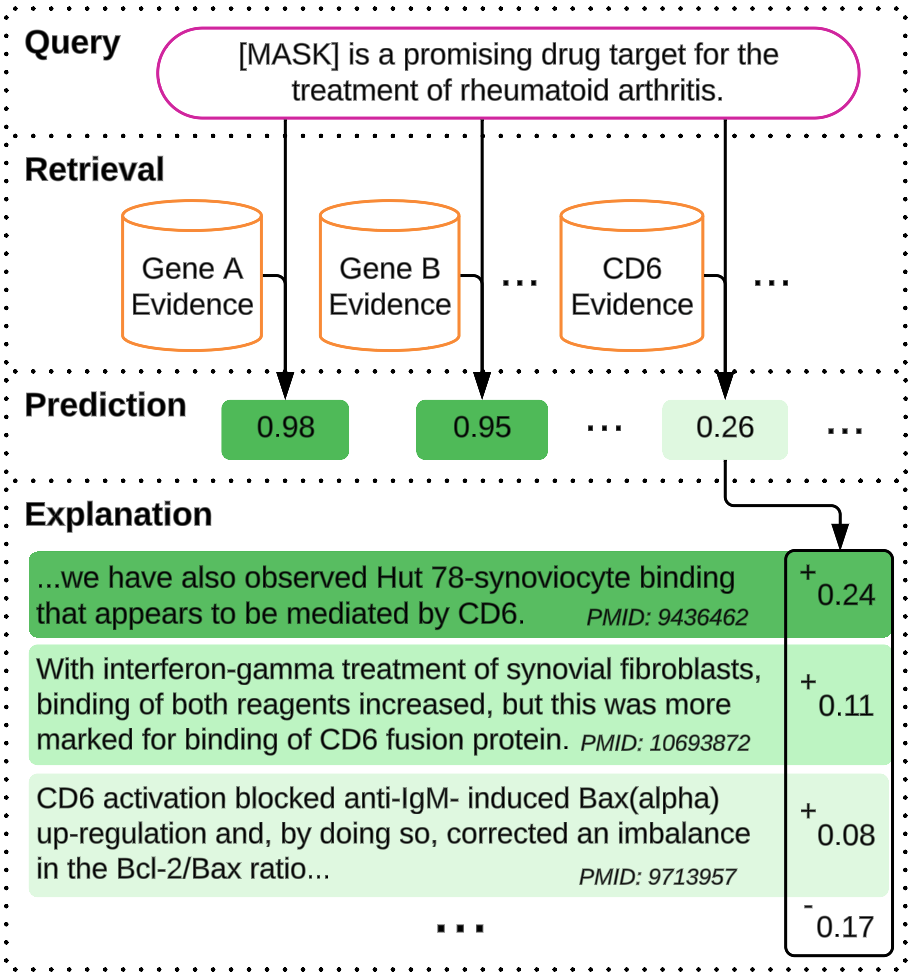}
\caption{\textbf{\acrshort{ours} drug target identification example}. \acrshort{ours} makes predictions based on retrieved evidence and provides explanations in terms of the evidence. \textbf{Query}: User queries are phrased in cloze-style, where [MASK] can be filled from a set of potential answers (named entities). For target identification, answers are the set of protein-coding genes (potential drug targets), and the query specifies a disease. \textbf{Retrieval}: \acrshort{ours} retrieves the evidence most relevant to the query for each potential answer, where evidence here is taken from across the biomedical literature that mentions the specific answer. \textbf{Prediction}: The model scores each answer based on the supporting evidence. \textbf{Explanation}: Each answer score is directly and quantitatively attributed to its retrieved evidence using Shapley values. Here, the best evidence is indirect, based on the role of CD6 in mechanisms central to rheumatoid arthritis pathology.}
\label{figure-easl-process}
\end{figure}

Beyond answering factual queries, a searchable knowledge base could provide evidence for queries without known answers, including scientific research questions (e.g. \emph{What are some promising drug targets to treat rheumatoid arthritis?}). By proposing new hypotheses supported by both direct and indirect scientific evidence, AI models could facilitate scientific discovery \cite{paliwal2020preclinical,aliper2023prediction,sourati2023accelerating}.

For high-stakes settings where acting on model hypotheses is costly or risky, an explainable model can mitigate risk by allowing a human expert to inspect the evidence and reasoning behind each prediction before acting on it (human-in-the-loop). Explainability can also help to identify model flaws or systemic biases, leading to improved performance and task alignment \cite{kulesza2015principles}.

Here, we introduce \acrfull{ours}, an approach for language model prediction with faithful and quantitative explanations (Figure \ref{figure-easl-process}). Given a cloze-style user query, \acrshort{ours} first retrieves the most relevant evidence from an evidence corpus, partitioned according to each possible answer. We consider a set of answers comprised of named entities. The model then scores each answer based on its supporting evidence to generate a ranked list. The \acrshort{ours} architecture represents potential answers explicitly in terms of their supporting evidence. In particular, the feature space is the evidence itself, enabling explainability with feature attribution methods to infer the contribution of each piece of evidence to the prediction. Here, we use Shapley values \cite{shapley1953value,lundberg2017unified}. In addition to explainability, we show that this evidence-oriented approach allows model predictions to be updated without retraining by modifying the corpus, such as introducing new evidence. Since \acrshort{ours} can generate a score for every answer in the answer set, it is particularly applicable in human-in-the-loop scenarios where many potential hypotheses are prioritized for user review.

With half of drugs failing to show efficacy when tested in human populations \cite{wong2019estimation}, often due to an ineffective choice of drug target, we developed \acrshort{ours} for drug target identification. Target identification is an especially challenging scientific discovery problem where specific genes or proteins (targets) are selected as the focus for developing new treatments, and where failures are extremely costly \cite{wouters2020}. We train \acrshort{ours} to score protein-coding genes by relevance to a user query based on a scientific literature corpus. We then augment the corpus with genetic associations by templating them into natural language, allowing the model to use both evidence sources. We show that Shapley values on individual pieces of evidence correlate with large language model (LLM) relevance assessments, which similarly correlate with human experts. Notably, when used to predict clinical trial outcomes, \acrshort{ours} significantly outperforms both genetics evidence, a widely recognised predictor in the pharmaceutical industry \cite{nelson2015support, trajanoska2023target}, and a few-shot, chain-of-thought, retrieval-augmentation GPT-4 baseline, a setup that in practice would also be prohibitively costly and sacrifices faithful explainability. \acrshort{ours} outperforms the genetics baseline even when supplied only with genetics evidence, suggesting that representing gene-trait associations in natural language improves generalization over a structured ontology. Finally, we show that \acrshort{ours}'s explainability enables the use of LLMs to audit prediction reasoning, further improving performance.

Alongside the clinical trial outcomes, we evaluate the model on two additional target identification benchmarks and make all three new benchmarks publicly available (Appendix \ref{appendix-eval-data-licenses}).

Our core contributions are as follows:
\begin{itemize}[leftmargin=0.6cm, itemsep=-0.5pt, topsep=-0.5pt]
    \item We introduce \acrshort{ours}, a novel architecture for retrieval-based high-stakes question answering, which scores the plausibility of each answer directly in terms of its supporting evidence, and thereby enables faithful, quantitative explainability using evidence-level Shapley values.
    \item We develop \acrshort{ours} for the challenging scientific discovery problem of drug target identification; it is not only as predictive of clinical trial outcomes as non-explainable literature-based baselines, but also surpasses a genetics approach used throughout the pharmaceutical industry.
    \item We release three new benchmarks to address the lack of publicly-available datasets for drug target identification and drive progress on this important scientific discovery problem.
\end{itemize}

\section{Related work}

\subsection{Language Models with Retrieval}

Many language models leverage retrieved text at inference time for question answering \cite{khandelwal2019generalization,karpukhin2020dense,guu2020retrieval,lewis2020retrieval,lee2020learning,izacard2021leveraging,borgeaud2022improving,izacard2022few}. \acrshort{ours} differs from these 
existing approaches by (1) scoring all possible answers in an answer set and (2) faithfully and quantitatively attributing each answer's score to evidence passages using Shapley values. This approach follows from the application: \acrshort{ours} is designed for answering research questions that merit deep user engagement (e.g. identifying potential drug targets for a disease) as opposed to typical factual recall tasks (e.g. identifying a country's capital city). Scoring many possible answers with faithful explanations allows a human to investigate them.

\acrshort{ours} perhaps bears the most resemblance to kNN-LM \cite{khandelwal2019generalization} which uses retrieval to improve next-token prediction. However, kNN-LM uses retrieval to augment a standard masked language model, while \acrshort{ours} is fully retrieval-based to enable evidence-driven explanations. The Fusion-in-Decoder (FiD) approach \cite{izacard2021leveraging} also bears a resemblance to \acrshort{ours}; both merge each piece of evidence with the query independently before jointly processing. FiD is motivated by efficiency and performance. We are additionally motivated by explainability. As discussed in depth in Appendix \ref{appendix-gpt-baseline}, faithfully explainable multi-label prediction with existing generative LLM architectures is largely infeasible.

\subsection{Explainability \& Data Attribution}

\acrshort{ours} is inspired by SHAP (SHapley Additive exPlanations) \cite{lundberg2017unified}, which explains model predictions by approximating feature-level Shapley values \cite{shapley1953value}. \acrshort{ours} extends feature attribution methods like SHAP to data, by using a retrieval-based architecture in which the feature space is comprised of evidence. \acrshort{ours} therefore also contrasts with explainability-focused training data attribution (TDA) methods \cite{hammoudeh2024training}, such as representer
point selection \cite{sui2021representer}, which evaluates the impact of training examples on predictions. Instead, \acrshort{ours} uses the evidence in the corpus at inference time for both prediction and explanation. Among TDA methods, Data Shapley  \cite{ghorbani2019data} also assigns Shapley values to data. Data Shapley focuses on explaining model performance rather than inference-time predictions.

SimplEx \cite{crabbe2021explaining} explains predictions by approximating an input in terms of a corpus of classified exemplars. SimplEx is general-purpose but indirect: the corpus illuminates black-box predictions, but does not impact them. In contrast, the corpus drives model predictions in \acrshort{ours}.

\subsection{Models for Hypothesis Generation}

The use of models in generating or evaluating scientific hypotheses is an emerging area of research. Knowledge graphs (KGs) are a popular approach for novel hypothesis generation, because their structure enables multi-hop inference between unconnected nodes. Novel hypotheses have been generated by subject-area experts directly querying and inspecting a KG \cite{smith2021expert}.

\citet{sourati2023accelerating} use KG patterns for material property prediction and drug re-purposing, additionally leveraging nodes for specific researchers to infer which discoveries are more or less likely to be discovered based on social dynamics. \citet{paliwal2020preclinical} used tensor factorization on a biomedical KG to predict future research findings and clinical trial outcomes for therapeutic drug targets. \citet{aliper2023prediction} similarly employed a biomedical KG to predict clinical trial outcomes; they used a graph transformer network ensembled with a tabular model leveraging clinical trial design features. \acrshort{ours} differs from these approaches by enabling explainability in terms of the evidence and operating directly on published research without needing to construct a KG.

In this vein, \citet{tshitoyan2019unsupervised} work with a materials science research corpus to identify new material properties. They use cosine similarity on unsupervised word embeddings, specifically word2vec \cite{mikolov2013distributed}. This resembles our parametric masked language model baseline, except that in our case embeddings are derived using a transformer. \citeauthor{tshitoyan2019unsupervised} suggest that word2vec enables indirect inference similar to that in a KG; for instance, a material never defined as thermoelectric may be mentioned alongside properties associated with thermoelectricity. We observe a similar phenomenon in \acrshort{ours}: for instance, a target never identified directly with a disease may still have been shown to regulate disease-relevant mechanisms (Figure \ref{figure-easl-process}) or to be genetically associated with relevant traits (Appendix \ref{appendix-cod-semantic-matching}). \acrshort{ours} can use these indirect findings as support.


\section{Methods}

\label{sec:methods}

We consider the problem of scoring $N$ potential answers \mbox{$\mathcal{A} = \{a_i\}_{i=1}^N$} to a user query $q$, to rank them from most to least relevant. To align with the training corpus (Section \ref{methods-mel-corpus}), we let $q$ be cloze-style (e.g. \textit{[MASK] is a promising drug target for the treatment of osteoporosis.}), where each answer $a_i$ represents a potential named entity at \textit{[MASK]}. \citet{lewis2019unsupervised} provides an approach to translate between cloze- and natural-style questions.

\begin{figure*}[t]
\begin{center}
\setlength{\belowcaptionskip}{-10pt}
\centerline{\includegraphics[width=\linewidth]{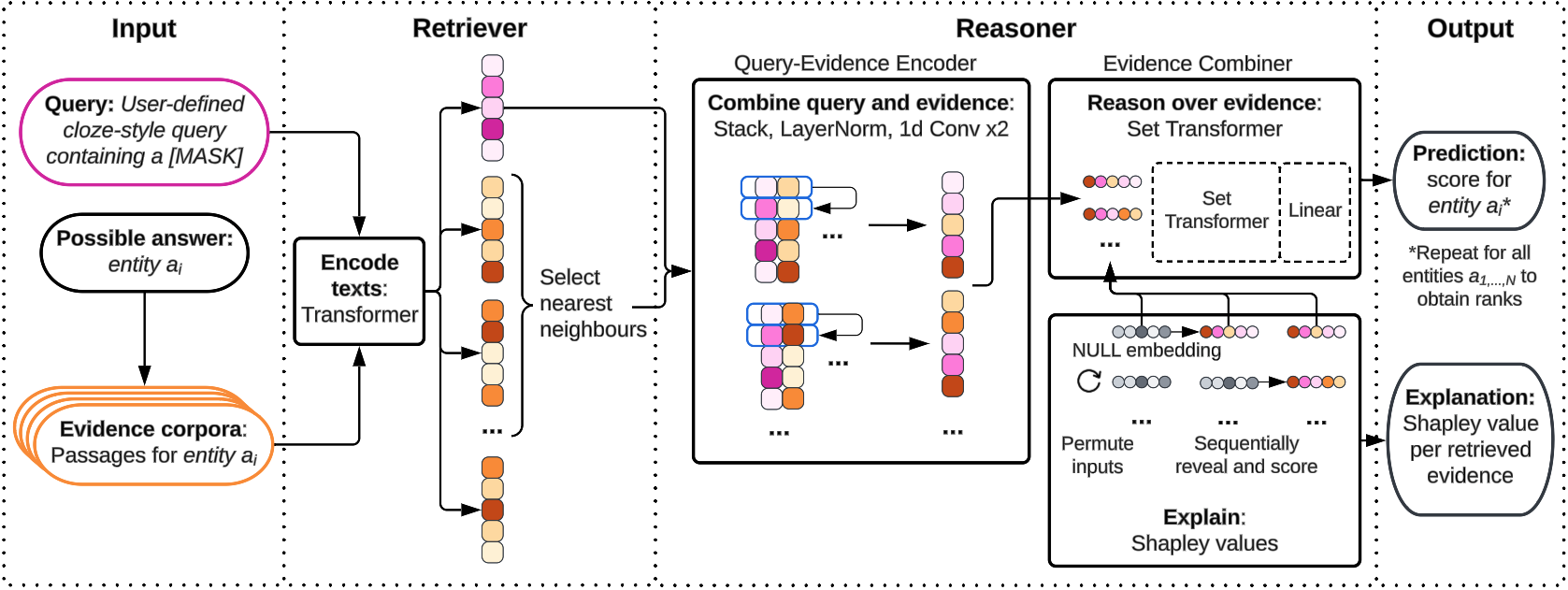}}
\caption{\textbf{\acrshort{ours} architecture schematic}. Illustration of \acrshort{ours} inference and explanation. \textbf{Input}: A user-defined cloze-style query, a possible answer (named entity) to evaluate, and a corpus of evidence passages corresponding to that answer entity with entity mentions replaced with \lbrack MASK\rbrack{}. \textbf{Retriever}: The query text is encoded with a transformer. All of the entity's evidence passages are encoded prior to inference, using the same encoder, and stored in a FAISS search index. The $k$ evidence passages with highest cosine similarity to the query are retrieved. \textbf{Reasoner}: Each evidence embedding is stacked with the query embedding. The resulting query-evidence pairs are layer-normalised before each pair is combined at corresponding dimensions into a single embedding using convolutional layers. All combined pair embeddings are passed to a set transformer, followed by a linear layer and sigmoid to obtain the binary probability. Shapley values for each pair (corresponding to each piece of evidence) can be computed to quantitatively explain the prediction. \textbf{Output}: To rank a set of answer entities $a_{1...N}$, binary probabilities are obtained independently for each. Shapley values attribute model predictions back to the evidence passages providing an explanation of the model's prediction.}
\label{figure-easl-model}
\end{center}
\end{figure*}

\subsection{Masked Entity-Linked Corpus}
\label{methods-mel-corpus}
Our approach uses a training corpus of textual passages, $\mathcal{D}$, each containing at least one named entity from the set of answer entities $\mathcal{A}$. Entity linking identifies and grounds entities in $\mathcal{A}$ in the corpus. For each passage, the span of every occurrence of a single entity is replaced by a \lbrack MASK\rbrack{} token. When the passage contains multiple unique entities in $\mathcal{A}$, we duplicate the passage with each masked in turn while the others appear as plain text. Each example is therefore a tuple $(a,d)$ consisting of an answer entity identifier $a \in \mathcal{A}$ and a masked text passage $d \in \mathcal{D}$ in which that entity occurs.

In application to drug target identification, $\mathcal{A}$ consisted of 19,176 protein-coding gene entities, hereafter referred to collectively as \textit{Genes}. $\mathcal{D}$ was an entity-linked corpus of 160 million sentences from scientific literature. For more details on the corpus and splits, including temporal splits to avoid leakage, see Appendix \ref{appendix-lit-dataset}; for entity linking see Appendix \ref{appendix-entity-linking}. $\mathcal{D}$ could in theory support other tasks (e.g. biomarker identification, drug repurposing, biological mechanism selection) by adjusting $\mathcal{A}$.

\subsection{Masked Language Model (MLM)}
\label{methods-mlm}

We first consider a parametric approach based on the pre-training method in \citet{brayne-etal-2022-masked}. We use an encoder-only transformer \cite{vaswani2017attention}, specifically a scaled-down version of PubMedBERT \cite{Gu_2021}. For query passages $d^q \in \mathcal{D}^{\mathrm{MLM}} \subset \mathcal{D}$ containing a masked answer $a_i \in \mathcal{A}$, we train to predict $p(a_i|d^q)$.

The query embedding is the mean over output embeddings at \lbrack MASK\rbrack{} token positions. We take the dot product with a learned embedding for each possible answer $a_i \in \mathcal{A}$, then apply a bias and softmax to predict $p(a_i|d^q) \, \forall \, a_i \in \mathcal{A}$. We train with cross-entropy loss. Pre-trained domain-specific model weights are available (e.g. PubMedBERT), but we train from scratch to avoid leakage from pre-training data in our temporally-split evaluations. 

This model is both a baseline (MLM) and the basis for the Retriever of \acrshort{ours} (Section \ref{methods-retriever-architecture}).

\subsection{\acrshort{ours} Retriever}
\label{methods-retriever-architecture}

We now consider our semi-parametric approach, \acrshort{ours}, which leverages retrieval from an evidence corpus. \acrshort{ours} combines a Retriever module and a Reasoner module (Figure \ref{figure-easl-model}). See Appendix \ref{appendix-easl-hyperparameters} for additional details of the \acrshort{ours} architecture, training and inference hyperparameters.

The MLM in Section \ref{methods-mlm} produces text embeddings that are trained to have a high inner-product with the paired answer embeddings in the answer set. We reasoned that two text embeddings would therefore have high similarity if they permit a similar distribution over answers, i.e. if they were semantically similar with respect to this task. This makes the MLM well-suited to identifying corpus passages that are relevant to the user query and so we used this MLM as the \acrshort{ours} Retriever.

We first used the MLM to embed each of the masked evidence passages in the evidence corpus $\mathcal{D}^e$, where $\mathcal{D}^e = \mathcal{D}^{\mathrm{MLM}}$ for Reasoner training (Section \ref{methods-reasoner}; typically $D^e = \mathcal{D}$ at inference). We partitioned evidence embeddings by the masked answer entity they contained, and created distinct FAISS search indices \cite{johnson2019billion} for each to enable efficient answer-specific retrieval. 

At inference time, the user's cloze-style query $q$ is encoded with the MLM. The Retriever selects $k$ evidence passages $[d^e_{1i}, ..., d^e_{ki}] \subset \mathcal{D}^e$ with the highest cosine similarity to $q$ from each answer $a_i$'s search index (we use $k=64$). The query embedding and retrieved evidence embeddings for each answer are the inputs to the Reasoner.

\subsection{\acrshort{ours} Reasoner}
\label{methods-reasoner}

\paragraph{Training Objective} We train the Reasoner with a binary cross entropy loss to differentiate positive examples ($L=1$) from negative examples ($L=0$), i.e.\ to learn $p(L=1|a_i,d^q)$ when taking an entity $a_i$ and masked query $d^q$ as input, where $d^q \in \mathcal{D}^q \subset \mathcal{D} \setminus \mathcal{D}^e$ ($\mathcal{D}^e$ excluded to avoid trivial inference by retrieving $d^q$ from $\mathcal{D}^e$). Positive examples were constructed from pairs $(a_p, d^q) \forall d^q \in \mathcal{D}^q$, where $a_p$ is the true masked answer in $d^q$. For each positive example, a corresponding negative example $(a_n, d^q)$ was constructed by uniformly sampling $a_n \in \mathcal{A} \backslash \{a_p\}$. For each $(a_i, d^q)$ pair, positive or negative, the Reasoner receives the top $k$ evidence passages $[d^e_{1i}, ..., d^e_{ki}]$ fetched by the Retriever from the retrieval corpus of $a_i$. For negatives, due to sampling of $a_n$, retrieved evidence corresponds to a different entity to the answer entity masked in the query. Under this sampling scheme, the objective $p(L=1|a_i,d^q)$ is closely related to the MLM multiclass objective $p(a_i|d^q)$ at optimality (Appendix \ref{appendix-objective-theory}); however, unlike multinomial regression, sampling avoids needing to retrieve evidence for all answers for each training example.

\paragraph{Inference} At inference time, we use $p(L=1|a_i,q)\, \forall \, a_i \in \mathcal{A}$ to score and rank the full answer set for the cloze-style query $q$, using the evidence fetched for $q$. This requires $|\mathcal{A}|$ nearest neighbour searches and forward passes through the Reasoner; however, since retrieval and reasoning independent for each answer, the process can be fully parallelized subject to computational resources. See Appendix \ref{appendix-speed} for profiling of inference speeds.

\paragraph{Architecture} The \acrshort{ours} Reasoner architecture is shown in Figure \ref{figure-easl-model}. First, the \textit{query-evidence encoder} $f: \mathbb{R}^h \times \mathbb{R}^h \rightarrow \mathbb{R}^h$  combines each of the $k$ evidence embeddings with the query independently. It stacks the evidence with the query to generate a tensor of size $[2,h]$; it then compresses the tensor into a vector of size $[1,h]$ using convolutional layers. The convolutional layers have a filter size of $[2,1]$ across each embedding dimension $h$, encoding the relationship between the query and evidence in each dimension. 

Next, the \textit{evidence combiner} $g: (\mathbb{R}^h)^k \rightarrow [0, 1]$ generates $p(L=1|a_i,q)$ from the $k$ query-evidence embeddings. There is no inherent ordering among the $k$ vectors, so we use a set transformer \cite{lee2019set}. 

Answers are masked in both the query and answer-specific evidence so that the Reasoner sees $a_i$ only indirectly via evidence embeddings. As a result, the score reflects the probability that query and evidence embeddings relate to the same entity.

\subsection{\acrshort{ours} Explanations}
\label{methods-shap}

\acrshort{ours} provides explanations in the form of Shapley values \cite{shapley1953value,lundberg2017unified} - the average expected marginal contribution of each piece of evidence to the overall model score for the query. Shapley values enable attribution of the model prediction back to pieces of retrieved evidence, such that they sum up to the overall score.

Multiple methods exist for rapidly approximating Shapley values on deep learning features \cite{lundberg2017unified}. Defining each of the $k$ inputs to the evidence combiner as a distinct feature gives a relatively small feature space, making it tractable to use a permutation sampling approach to approximate Shapley values. See Appendix \ref{appendix-shap} for the full algorithm and Appendix \ref{appendix-speed} for profiling.

During training, we replaced query-evidence features at random with a learned NULL embedding. In addition to acting as a regularizer (akin to dropout), introducing the NULL embedding during training ensured that the model could handle missing features robustly when estimating Shapley values. For each training example, the evidence dropout rate was sampled in \( \text{Uniform}(0, 1) \).

\subsection{Post-hoc Frequency Bias Correction}
\label{methods-post-hoc-correction}

Many answer sets suffer from class imbalance. In drug target identification, some targets are significantly more well-studied than others. As a result, the learned model $p(a_i|q)$ can be strongly correlated with the prior $p(a_i)$. 

While such bias can be informative (e.g. reflecting the fact that some targets are involved in more diseases than others) it can also be misleading (e.g. reflecting publishing trends rather than underlying biology). To flexibly control for bias, we introduce a method to correct the model output score based on the frequency of answers in the training corpus, resulting in an up-ranking of less frequently mentioned answers, as detailed in Appendix \ref{appendix-bias-corr-theory}. The correction is parameterized by $c \in [0, 1]$: when $c=0$ the scores and rankings are unaltered; when $c=1$, the rankings reflect the pointwise mutual information (PMI) of the query and answer, inspired by PMI use in NLP co-occurrence statistics \cite{church1990word}. In the results we report both uncorrected ($c=0$; \acrshort{ours}-uncor) and partially corrected ($c=0.5$; \acrshort{ours}-cor; selected using validation set, Appendix \ref{appendix-easl-hyperparameters}) rankings. In Shapley value explanations, the bias correction can be represented as an additive feature.


\section{Experiments and Results}
\label{section-experiments-and-results}

We evaluate \acrshort{ours} performance on three datasets aligned with drug target identification, which we publicly release (Appendix \ref{appendix-eval-data-licenses}):

\begin{itemize}[leftmargin=0.6cm, itemsep=-0.5pt]
    \item \textbf{Held-out Biomedical Literature}: Predicting masked genes in biomedical literature sentences from abstracts published after the publication of the training data and retrieval corpus.
    \item \textbf{Gene Description Facts}: Predicting masked genes in sentences from human-curated gene descriptions adapted from UniProt \cite{uniprot2022}.
    \item \textbf{Clinical Trial Outcomes}: Retrospectively predicting success or failure in clinical trials based on evidence published before the trials, using the disease indication and drug target (gene). 
\end{itemize}

For \textit{Gene Description Facts} and \textit{Clinical Trial Outcomes}, we also construct \textit{Evidence Annotations} datasets to evaluate the alignment of \acrshort{ours} explanations with expert reasoning. We look at the strength of relationship between \acrshort{ours} Shapley values and GPT-4 \cite{achiam2023gpt} binary annotations of whether each piece of explanatory evidence is relevant or irrelevant to the query. We validate GPT-4 annotations against human expert annotations. More detailed usability testing of \acrshort{ours} Shapley values is left to future work.

For dataset summary statistics see Appendix \ref{appendix-eval-data-sizes}.

\subsection{Metrics}
\label{methods-metrics}

For ranking \textit{Genes} on \textit{Held-out Biomedical Literature} and \textit{Gene Description Facts}, we report mean reciprocal rank (MRR), mean rank (MR), hits@10 (h@10) and hits@200 (h@200). For \textit{Gene Description Facts}, we used macro metrics to give each gene equal weight irrespective of frequency. For \textit{Clinical Trial Outcomes} we report AUROC, and include relative success results in Appendix \ref{appendix-cod-results} for consistency with \citet{minikel2023refining}. We compare AUROCs using DeLong test, and relative successes using Z-test, reporting confidence intervals using Katz method \cite{katz1978}. For \textit{Evidence Annotations}, we report AUROC for the \acrshort{ours} Shapley scores of evidence sentences against GPT-4 annotations, and accuracy when validating GPT-4 against human expert annotations.

\subsection{Baselines and Ablations}
\label{results-baselines-ablations}

In addition to MLM (Section \ref{methods-mlm}), we include two baselines throughout: FREQ and MCS. For FREQ, entities were scored according to their frequency in the training set of $\mathcal{D}$. For MCS (mean cosine similarity), each entity $a_i$ was scored by computing $\frac{1}{64} \sum_{j=1}^{64} (d^e_{ji} \cdot q)/(\|d^e_{ji}\| \|q\|)$ for the query $q$. 

For \textit{Clinical Trial Outcomes}, we include a competitive genetics baseline used throughout the pharmaceutical industry (in-depth setup in Appendix \ref{appendix-cod-genetics-baseline}). Alongside other relative success results in Appendix \ref{appendix-cod-results}, we compare to a few-shot, chain-of-thought, retrieval-augmented GPT-4 baseline (setup in Appendix \ref{appendix-gpt-baseline}). For extensive ablations of \acrshort{ours}, including the Retriever, Reasoner and literature bias correction, see Appendix \ref{appendix-ablations}.

\subsection{Held-out biomedical literature}

Given their greater orthogonality to the \acrshort{ours} training objective, we choose to focus on \textit{Gene Description Facts} and \textit{Clinical Trial Outcomes} in the main text, and save complete results for \textit{Held-out Biomedical Literature} for Appendix \ref{appendix-bml} (Table \ref{literature-table}). For the latter, \acrshort{ours} outperformed all baselines and was able to leverage retrieved literature that it was not trained on, further improving performance.

\subsection{Gene Description Facts}
\label{sec:gdf}

\paragraph{Dataset Construction} We sought to validate that \acrshort{ours} could perform well on predicting protein-coding genes in human-curated descriptions of gene function. We extracted descriptions of protein functions for our \textit{Genes} entities from \href{https://www.uniprot.org/}{UniProt (Universal Protein Resource)} \cite{uniprot2022}. Each UniProt description is a human-written summary of a protein's function, and consists of one or more sentences. We used a combination of regular expressions and Anthropic's Claude 2.0 to extract \lbrack MASK\rbrack{}-containing facts from each description. Further details of the source and preprocessing of the dataset, including the Claude prompt and an example gene description with extracted facts, are found in Appendix \ref{appendix-gene-descriptions-creation}. \acrshort{ours} was trained on, and retrieved from, all years of literature evidence for the \textit{Gene Description Facts} evaluation. 

We also constructed an \textit{Evidence Annotations} dataset by having GPT-4 (prompt in Appendix \ref{appendix-gdf-evidence-annotations}) annotate as query-relevant or irrelevant, all evidence for 50 randomly sampled \textit{Gene Description Facts} query-entity pairs (positive examples), and the same 50 queries with randomly sampled alternative entities (negative examples), obtaining 6400 annotated query-evidence pairs. To validate GPT-4 annotations, a human drug discovery expert following the GPT-4 prompt annotated all 512 query-evidence pairs for a subset of 8 randomly sampled examples (4 positive, 4 negative).

\paragraph{Results} \acrshort{ours} substantially improved on all baselines, both with and without bias correction (Table \ref{gene-descriptions-table}). As expected, bias correction was helpful. \acrshort{ours} metrics here appear to show greater improvement over baselines than for the \textit{Held-out Biomedical Literature} dataset in Table \ref{literature-table}. This may reflect a tendency for gene descriptions to describe more well-established knowledge than literature; as a result, \acrshort{ours} may benefit from its access to such facts, when more directly stated in the retrieved evidence sentences.

Additionally, there was a strong correlation between evidence Shapley values and GPT-4 relevance annotations (AUROC: 0.824). See Appendix \ref{appendix-gdf-evidence-annotations-examples} for a case study of examples. Combined with a $71.5$\% agreement rate between GPT-4 and human-expert annotations, the agreement between \acrshort{ours} and GPT-4 suggests that \acrshort{ours} has correctly learnt to prioritise evidence for its predictions.

\begin{table}[t]
\caption{\textbf{Gene Description Facts}: \acrshort{ours} macro ranking metrics.}
\label{gene-descriptions-table}
\vskip 0.15in
\begin{center}
\begin{small}
\begin{sc}
\vspace{-8pt}
\begin{tabular}{l|ccc|cc}
\toprule
Metric & \multicolumn{3}{c|}{Baselines} & \multicolumn{2}{c}{\acrshort{ours}} \\
       & Freq & MCS & MLM & Uncor & Cor \\
\midrule
MRR    & $<$0.001 & 0.176 & 0.167 & 0.202 & \textbf{0.260} \\
MR     & 8252 & 1776 & 2208 & 937 & \textbf{599} \\
h@10 & $<$0.001 & 0.309 & 0.296 & 0.349 & \textbf{0.434} \\
h@200 & 0.013 & 0.622 & 0.590 & 0.701 & \textbf{0.776} \\
\bottomrule
\end{tabular}
\vspace{-7pt}
\end{sc}
\end{small}
\end{center}
\end{table}

\subsection{Clinical Trial Outcomes}
\label{sec:cod}

\paragraph{Dataset Construction} We constructed a benchmark of gene-disease pairs (therapeutic hypotheses) from clinical trials as per \citealt{nelson2015support}, using the PharmaProjects database \cite{pharmaprojects} (1,449 success, 4,222 failure, Appendix \ref{appendix-cod-creation}). This benchmark focused on \textit{in vivo} efficacy of therapeutic hypotheses as demonstrated by the transition of drugs associated with such hypotheses from Phase II/III clinical trials to regulatory approval.

To avoid leakage due to reporting of clinical trials in the literature, we removed drugs investigated prior to 2005 (Appendix \ref{appendix-cod-creation}) and used pre-2005 literature for \acrshort{ours} training and retrieval  (Appendix \ref{appendix-lit-dataset}). We scored therapeutic hypotheses using a query template ``\textit{\lbrack MASK\rbrack{} is a promising drug target for the treatment of \{DISEASE\}.}", substituting in the PharmaProjects disease (Appendix \ref{appendix-cleaning-mesh-terms}).

As the ability of genetics methods such as locus-to-gene \cite{mountjoy2021} to predict successful clinical development \cite{nelson2015support, ochoa2022human, minikel2023refining} drives their wide use in target identification, we used the most recently published PharmaProjects-aligned dataset of genetics predictions \cite{minikel2023refining} (Appendix \ref{appendix-cod-genetics-baseline}) as a competitive baseline. In order to validate our \textit{Clinical Trial Outcomes} data, we corroborated the published result \cite{minikel2023refining} that the probability of clinical success of therapeutic hypotheses supported by genetics evidence is approximately double the probability without supporting genetics evidence (relative success: 1.98; 95\% CI (1.76, 2.24); Appendix \ref{appendix-cod-genetic-insight}).

We also constructed an \textit{Evidence Annotations} dataset with GPT-4 (prompt in Appendix \ref{appendix-cod-evidence-annotations}) assessing the relevance of all 64 evidence passages for 100 \textit{Clinical Trial Outcome} therapeutic hypotheses (50 success, 50 failure; randomly sampled), obtaining 6400 annotated query-evidence pairs. To validate GPT-4 annotations, a human drug discovery expert following the GPT-4 prompt annotated all 512 query-evidence pairs for a subset of 8 hypotheses for which they had most knowledge (4 success, 4 failure).

\paragraph{Multimodality via Templating into Natural Language} We assessed \acrshort{ours}'s ability to reason from genetics by generating a sentence for every row in the genetics dataset used in the genetics baseline (77,645 total), with the simple template ``\textit{\lbrack MASK\rbrack{} is genetically associated with \{MeSH name\}.}''. The MeSH name, as supplied in  \citealt{minikel2023refining}, was programmatically reformatted to better align with naming conventions in the biomedical literature (details in Appendix \ref{appendix-cleaning-mesh-terms}). This genetics corpus was given to the \acrshort{ours} Retriever alone and in combination with the pre-2005 biomedical literature.

\begin{table}
\caption{\textbf{Clinical Trial Outcomes}: AUROC for \acrshort{ours} with retrieval corpus of literature-alone, genetics-alone, or both combined. For relative success metrics, including comparison to a few-shot, chain-of-thought, retrieval-augmented GPT-4 baseline, see Figure \ref{appendix-figure-relative-success} and Appendix \ref{appendix-cod-results}.}
\label{cod-table-1}
\begin{center}
\begin{small}
\begin{sc}
\vspace{-2pt}
\begin{tabular}{ll|cc}
\toprule
 Model & Corpus &  AUROC \\
\midrule
Genetic         & Genetics & 0.545 \\
Freq        & Literature & 0.561 \\
MCS         & Literature & 0.623  \\
MLM         & Literature &  0.630 \\
\hline
\acrshort{ours}-uncor & Genetics & 0.579 \\
\acrshort{ours}-uncor & Literature & 0.629 \\
\acrshort{ours}-cor & Literature & 0.632 \\
\acrshort{ours}-cor & Both & \textbf{0.633}\\
\midrule
\midrule
\acrshort{ours}-audit & Both & \textbf{0.638} \\ 
\bottomrule
\end{tabular}
\vspace{-6pt}
\end{sc}
\end{small}
\end{center}
\end{table}

\paragraph{Results} Table \ref{cod-table-1} shows primary results, while Appendix \ref{appendix-cod-results} includes several further results and detailed discussions, including on relative success (Appendices \ref{appendix-cod-rs}-\ref{appendix-cod-all-diseases}; Figure \ref{appendix-figure-relative-success}). Overall, \acrshort{ours} variants incorporating biomedical literature matched or outperformed all baselines.

Notably, \acrshort{ours} significantly outperformed the widely-used genetics baseline (Genetic) when leveraging only the exact same underlying genetics data templated into sentences (\acrshort{ours}-uncor (genetic); $p<0.001$). This could be explained by the language model's capacity to leverage ``soft" semantic associations (e.g. recognizing correlations between diseases / traits beyond ontological similarity), as corroborated by the inspection of high-scoring genetics evidence (Appendix \ref{appendix-cod-semantic-matching}; Figure \ref{appendix-figure-trait-trait-similarity}). The addition of literature resulted in a significant further improvement ($p<0.001$). The relative under-performance of models using genetics data alone compared to those using biomedical literature likely reflects the lack of genetic coverage of diseases, despite it being predictive when available. In contrast, the literature has broad coverage across diseases. Figure \ref{appendix-figure-disease-area-auc} (Appendix \ref{appendix-cod-by-disease-area}) shows performance by disease area with greater variability for genetics.

\acrshort{ours} also significantly outperformed the few-shot, chain-of-thought prompted GPT-4 baseline with retrieval augmentation. The full method and results for this baseline are described in Appendix \ref{appendix-gpt-baseline} and \ref{appendix-cod-gpt-baseline-results} respectively.

There was only a marginal improvement from combining templated genetics evidence and the biomedical literature over literature alone. This could be explained by the $\sim$200:1 balance of literature to genetics-derived sentences in the evidence corpus, and the potential redundancy of the genetics evidence given information already represented in the literature. Additional approaches to combining data sources, with similar performance, are compared in Appendix \ref{appendix-cod-combined-retrieval-approaches} (Table \ref{cod-table-multimodality-method}).

Evidence Shapley values correlated with binary GPT-4 relevance annotations (AUROC: 0.665) and GPT-4 with human-expert annotations ($82.2$\% agreement rate). Together, this suggests moderate agreement on evidence relevance. See Appendix \ref{appendix-cod-evidence-annotations-examples} for a case study of examples.

\subsection{Auditing Explanation Evidence}
\label{results-auditing}

We sought to assess the hypothesis that \acrshort{ours} explanations could enable human- or LLM-in-the-loop feedback to remove false positive evidence. Pooling \acrshort{ours} predictions on the \emph{Clinical Trial Outcomes} dataset, we used GPT-4 to annotate the relevance of 20,000 query-evidence pairs with the highest Shapley values (computed on pre-sigmoid outputs). We then reran \acrshort{ours}-cor inference on the full dataset, replacing evidence labelled as irrelevant with the NULL embedding, yielding a small but significant improvement (\acrshort{ours}-audit, Table \ref{cod-table-1}, $p=0.004$). Said differently, by allowing evidence to be audited, \acrshort{ours}'s explainability enabled further performance improvement. For the GPT-4 prompt, and auditing examples, see Appendices \ref{appendix-cod-evidence-annotations} and \ref{appendix-cod-auditing-examples}.


\section{Conclusions}

By retrieving evidence to make predictions, \acrshort{ours} enables faithful and quantitative explainability, a critical feature in complex, high-stakes settings such as drug target identification. \acrshort{ours} matched or outperformed all baselines across the three target identification evaluation tasks. Combined with the proposed bias correction technique, this improves the ability to make informed predictions about novel and understudied, but promising targets. Finally, \acrshort{ours} outperformed a widely-used competing approach on the important and challenging task of predicting clinical trial efficacy outcomes, without task-specific fine-tuning. Performance was further improved by auditing \acrshort{ours} explanations using GPT-4, an approach made possible by the retrieval-based setup. We show here that retrieval can provide not only performance and flexibility advantages, but also significantly improved transparency into how the model reasons from evidence.


\section{Limitations}
\label{sec:lim-future}

Retrieving evidence at inference time to make predictions has a cost: each answer score requires a vector search over the answer's evidence, followed by a model forward pass. In comparison, predicting with a multiclass model (MLM) requires a single forward pass without retrieval. For efficient scaling, retrieval and reasoning can be parallelized across answers (Appendix \ref{appendix-speed}).

Retrieval-based inference has flexibility benefits beyond those explored here. By filtering retrieved evidence on document metadata, users could customize the ranking at inference time; with a scientific literature dataset, this could include filtering supporting evidence to specific timespans, publications, impact factors, paper sections, or keywords.

The performance of a retrieval-based approach is expected to be sensitive to the completeness of the underlying corpus. However, R2E explanations help to make limitations or biases of the corpus more transparent than would be the case for a fully parametric approach, and parametric approaches are also sensitive to their training corpus.

In Sections \ref{sec:gdf} and \ref{sec:cod}, we applied the model directly to downstream tasks; in the case of clinical trials, we simply adopted a one-size-fits-all query template. Instead, the system could be fine-tuned for the task of interest. Fine-tuning with human feedback is of particular interest here, since with \acrshort{ours} a user can focus on faulty \textit{evidence use} (as opposed to a faulty prediction). Similarly, an LLM could be used to generate evidence-level labels for model fine-tuning in addition to the inference-time auditing described in Section \ref{results-auditing}. 

The evidence templating approach used for genetics in Section \ref{sec:cod} is relatively general, and could be applied to other data modalities, such as transcriptomics evidence in drug discovery. We focused on genetics because it is well-established as being predictive of clinical trial outcomes. For new modalities, care should be taken with respect to the distribution of the training data. For example, for scientific applications, evidence should be templated consistently with how it might be discussed in the literature corpus.

Performance gains might be made by scaling the Retriever and Reasoner, as well as extending to longer literature passages to increase context, for example paragraphs instead of sentences.

\section{Ethical Considerations}
\label{sec:impact}

As detailed in Section \ref{intro}, the explainability of \acrshort{ours} has the potential to positively impact the utility and adoption of models in high-stakes human-in-the-loop settings where explainability is often paramount, as exemplified by target identification. For target identification specifically, the improvements here could have significant positive consequences for the success of drug development programs and therefore the rate at which new more efficacious therapies become available to patients.

The application of \acrshort{ours} to predict and explain protein-coding genes in response to a user query is quite different to either the generality of large language models or the structural biology and chemistry foci of the AI-enabled biological tools most typically associated with any potential dual risk concern. As with other tools that facilitate biomedical research and understanding, the ability to identify and understand particular genes could be applied in a range of use cases. For this paper, we do not believe there to be material risks to highlight, especially noting: (1) We are not releasing proprietary training data, code, or model weights; (2) Explanations provided by \acrshort{ours} are either publicly-available extracts from the scientific literature or non-textual data templated in natural language, and can be interpreted by expert users in the context of their wider biomedical understanding, but do not significantly lower the barrier to entry for non-experts users; (3) \acrshort{ours} is predicting at the level of drug targets, with multiple complex downstream steps required to translate the identification of a target that may achieve a particular biological effect, into a capability to intervene on that target.


\section*{Acknowledgments}

The authors would like to thank Nicola Richmond and Julien Fauqueur for their helpful comments and feedback during drafting, Bradleigh Whitton and James Grey for applying their biological expertise and consenting to the use of their evidence annotations for validation of GPT-4 \textit{Evidence Annotations} datasets (Clinical Trial Outcomes and Gene Description Facts respectively), Alison McGarvey and Eryk Kropiwnicki for their help and advice on the process for templating genetics data into sentences, Antonios Poulakakis Daktylidis for their assistance with validation of the \textit{Clinical Trial Outcomes} data, and Hao-Chih Lee for their helpful insights on evaluations of evidence explanations.

\bibliography{r2e}

\appendix
\onecolumn

\section{Accessing Evaluation Datasets}
\label{appendix-eval-data-licenses}

We make the three performance evaluation datasets used in this paper publicly available as part of the Supplementary Material, licensed under \href{http://creativecommons.org/licenses/by-nc-sa/4.0/}{CC BY-NC-SA 4.0}. Specific licensing information for the datasets is as follows:
\begin{itemize}
    \item \textit{Clinical Trials Outcomes} is licensed under \href{http://creativecommons.org/licenses/by-nc-sa/4.0/}{CC BY-NC-SA 4.0}. We have permission from Citeline PharmaProjects to publicly release the subset of their data that is used here.
    \item \textit{Gene Description Facts} is licensed under \href{http://creativecommons.org/licenses/by-nc-sa/4.0/}{CC BY-NC-SA 4.0}. It is adapted from "Universal Protein Resource (UniProt)" by Uniprot Consortium, used under \href{https://creativecommons.org/licenses/by/4.0/}{CC BY 4.0}.
    \item \textit{Held-out Biomedical Literature} validation and test dataset sentences are courtesy of the National Library of Medicine.
\end{itemize}

We make the three performance evaluation datasets used in this paper (see Section \ref{section-experiments-and-results}) publicly available, licensed under \href{http://creativecommons.org/licenses/by-nc-sa/4.0/}{CC BY-NC-SA 4.0}, at: \href{https://github.com/BenevolentAI/r2e-evaluation-data}{https://github.com/BenevolentAI/r2e-evaluation-data}. Specific licensing information for the datasets is as follows:
\begin{itemize}
    \item \textit{Clinical Trials Outcomes} \copyright{} 2024 by \href{https://www.benevolent.com/}{BenevolentAI} is licensed under \href{http://creativecommons.org/licenses/by-nc-sa/4.0/}{CC BY-NC-SA 4.0}. We have permission from Citeline PharmaProjects to publicly release the subset of their data that is used here.
    \item \textit{Gene Description Facts} \copyright{} 2024 by \href{https://www.benevolent.com/}{BenevolentAI} is licensed under \href{http://creativecommons.org/licenses/by-nc-sa/4.0/}{CC BY-NC-SA 4.0}. It is adapted from "Universal Protein Resource (UniProt)" by Uniprot Consortium, used under \href{https://creativecommons.org/licenses/by/4.0/}{CC BY 4.0}.
    \item \textit{Held-out Biomedical Literature} validation and test dataset sentences are courtesy of the National Library of Medicine.
\end{itemize}

\section{Masked Entity-Linked Corpus, Dataset Splits \& Sizes}
\label{appendix-lit-dataset}

The large-scale corpus of scientific documents consisted of open access PubMed abstracts and PMC full texts as well as paid access Springer, Wiley and Elsevier full texts. We performed entity linking using a proprietary method (Appendix \ref{appendix-entity-linking}), however any entity linking approach may be used (e.g. dictionary matching). Individual sentences were used as passages.

We filtered to sentences in the corpus that contained both: i) one or more protein-coding genes (entity set referred to as \textit{Genes}), and ii) one or more non-gene grounded biomedical entities (e.g. diseases, biological pathways etc.), to select for an informative corpus. This process yielded 160 million sentences.

\begin{figure*}[ht!]
\setlength{\belowcaptionskip}{-8pt}
\begin{center}
\centerline{\includegraphics[width=\linewidth]{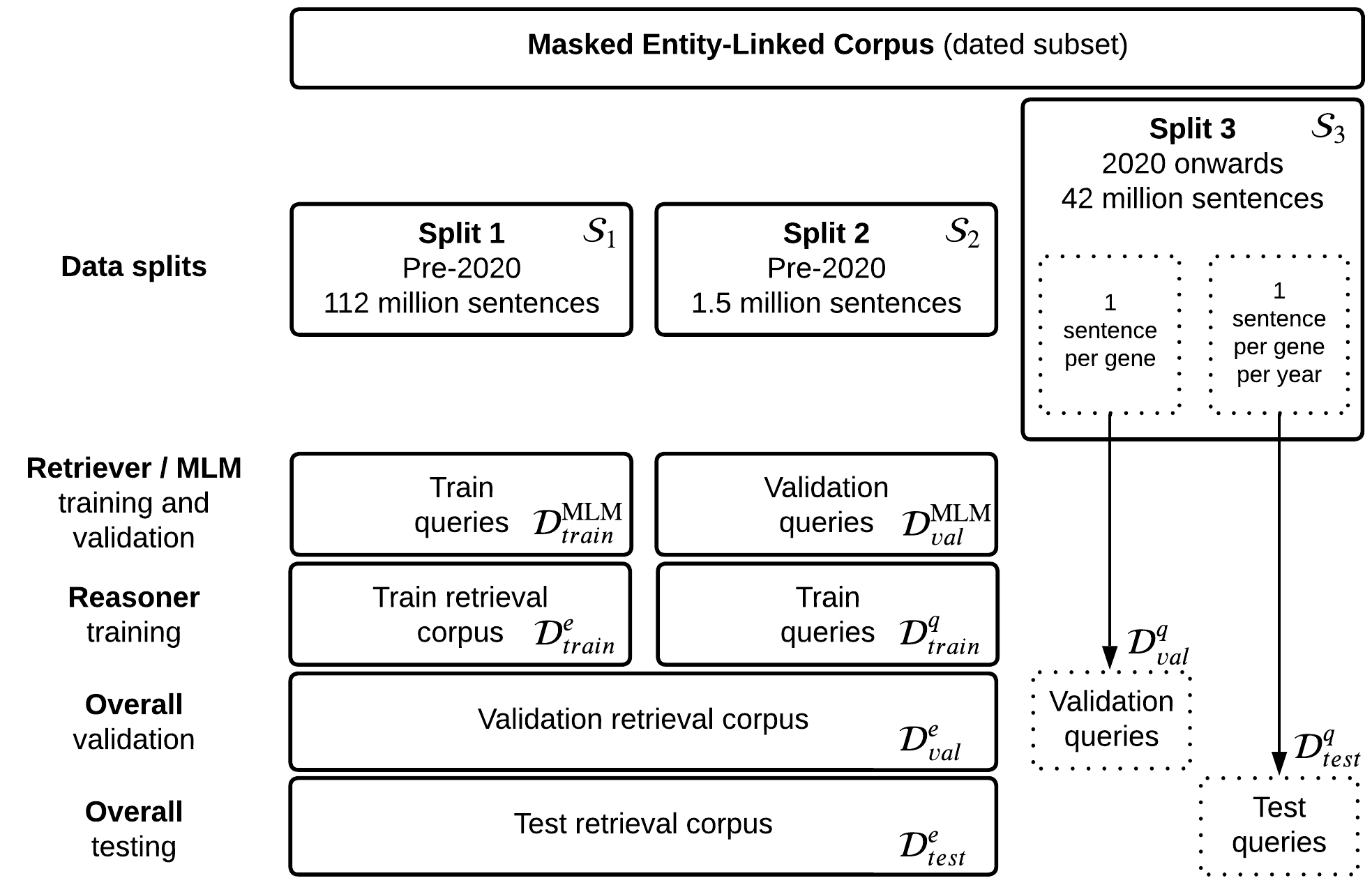}}
\caption{\textbf{\textit{Masked entity-linked corpus} for \textit{Held-out Biomedical Literature} experiments}. Here we illustrate how the \textit{masked entity-linked corpus} was partitioned to enable Reasoner/MLM and Retriever training, validation, and testing. Specifically the example of a 2020 year split setup is shown, as was used for \textit{Held-out Biomedical Literature} experiments.}
\label{figure-data-splits}
\end{center}
\end{figure*}

We created three distinct corpus splits $\mathcal{S}_1$, $\mathcal{S}_2$, and $\mathcal{S}_3$ (Figure \ref{figure-data-splits}). These splits were generated at the level of entire documents to reduce the occurrence of highly similar sentences between splits.

For \textit{Held-out Biomedical Literature} (Appendix \ref{appendix-bml}) and \textit{Clinical Trial Outcomes} (Section \ref{sec:cod}) experiments, where evaluation queries were associated with metadata for year of publication and earliest clinical development date respectively, a temporal year split setup was used to ensure models trained on and retrieved from sentences prior to the start year of the evaluation data. Specifically, for these year split experiments, $\mathcal{S}_1$ and $\mathcal{S}_2$ were random samples from \textit{before} the split year with 1.5 million sentences allocated to $\mathcal{S}_2$ and the remainder to $\mathcal{S}_1$. $\mathcal{S}_3$ contained all sentences from documents \textit{after} the split year. A split year of 2005 was used for \textit{Clinical Trial Outcomes} ($|\mathcal{S}_1| = $ 16.2 million sentences), and a split year of 2020 for \textit{Held-out Biomedical Literature} ($|\mathcal{S}_1| = $ 112 million sentences).

For \textit{Gene Description Facts} experiments (Section \ref{sec:gdf}), where evaluation queries did not correspond to a particular year, no year split was used. Specifically, $\mathcal{S}_1$, $\mathcal{S}_2$, and $\mathcal{S}_3$ were all random samples of the corpus, with 1.5 million sentences allocated to each of $\mathcal{S}_2$ and $\mathcal{S}_3$, and the remainder to $\mathcal{S}_1$ (157 million sentences).

Training, validation and testing datasets were then constructed for both \acrshort{ours} Retriever / MLM and \acrshort{ours} Reasoner, by using the appropriate $\mathcal{S}_1$, $\mathcal{S}_2$, and $\mathcal{S}_3$ splits. 

For the \acrshort{ours} Retriever / MLM, training and validation datasets were composed as follows:

\begin{itemize}
    \item $\mathcal{D}^{\mathrm{MLM}}_{train} = \mathcal{S}_1$
    \item $\mathcal{D}^{\mathrm{MLM}}_{val} = \mathcal{S}_2$
\end{itemize}

For the \acrshort{ours} Reasoner, for each of train, validation and test, both retrieval and query corpora were needed, to ensure query sentences were not also included in the retrieval corpus. We use $\mathcal{D}^e$ to refer to a retrieval corpus of evidence sentences and $\mathcal{D}^q$ to refer to the query corpus of sentences. The datasets were composed as follows: 

\begin{itemize}
    \item $\mathcal{D}^{e}_{train} = \mathcal{S}_1$
    \item $\mathcal{D}^{q}_{train} = \mathcal{S}_2$
    \item $\mathcal{D}^{e}_{val} = \mathcal{S}_1 \cup \mathcal{S}_2$
    \item $\mathcal{D}^{q}_{val} \subset \mathcal{S}_3$
    \item $\mathcal{D}^{e}_{test} = \mathcal{S}_1 \cup \mathcal{S}_2$
    \item $\mathcal{D}^{q}_{test} \subset \mathcal{S}_3 : \mathcal{D}^{q}_{test} \cap \mathcal{D}^{q}_{val} = \emptyset$, i.e. a held-out subset of $\mathcal{S}_3$, without overlap with $\mathcal{D}^{q}_{val}$
\end{itemize}

The above splitting procedure is illustrated in Figure \ref{figure-data-splits} for the case of the 2020 year split setup used for \textit{Held-out Biomedical Literature} experiments. For this \textit{Held-out Biomedical Literature} setup, the disjoint subsets sampled from $\mathcal{S}_3$ and used to create overall validation ($\mathcal{D}^{q}_{val}$) and test ($\mathcal{D}^{q}_{test}$) queries, are those used to report ranking metric evaluations over all genes in \textit{Genes}; namely the:
\begin{itemize}
    \item \textit{Held-out Biomedical Literature} validation dataset: 1 sentence per gene, sampled from publicly-available abstract section sentences from 2020 onwards. Used for hyperparameter selection and ablation experiments described in Appendices \ref{appendix-easl-hyperparameters} \& \ref{appendix-ablations} respectively.
    \item \textit{Held-out Biomedical Literature} test dataset: 1 sentence per gene per year for 2020 onwards, sampled from publicly-available abstract section sentences. Used for evaluations described in Section \ref{section-experiments-and-results} and Appendix \ref{appendix-bml}, including evaluation of the MLM and other baselines.
\end{itemize}

Note the key difference between this 2020 year split setup for \textit{Held-out Biomedical Literature}, and the setups for the other two evaluation datasets were:
\begin{itemize}
    \item Different year splits (as described above)
    \item The queries used in evaluation were derived from those specific evaluation datasets, not a held-out split of the literature corpus (i.e. ${\mathcal{D}}^{q}_{eval} \neq {\mathcal{D}}^{q}_{test}$)
\end{itemize}

\section{Entity Linking}
\label{appendix-entity-linking}
We used a proprietary entity linking methodology based on dictionaries of entities and synonyms, as well as an abbreviation detection algorithm and a model that resolves type ambiguities based on the context of each mention. The dictionaries were created from several sources.

\begin{enumerate}
    \item External ontologies.
    \item Human annotations of synonyms discovered by machine learning methods.
    \item Automatic synonym generation to cover e.g. variation in punctuation, Greek letters and plurals of terms.
\end{enumerate}

For the protein-coding gene target entities, referred to as \textit{Genes} and used throughout the paper, we ground both gene and protein forms to the same entity, under the assumption of a 1:1 relationship between a gene and the protein it encodes.

\section{\acrshort{ours} Hyperparameters}
\label{appendix-easl-hyperparameters}

The \acrshort{ours} model was implemented using PyTorch deep learning library \cite{paszke2019pytorch}.

All sentences were tokenized, and then truncated and padded to a length of 128, using the same vocabulary as PubMedBERT \cite{Gu_2021}. Pre-processing of training examples for both Retriever and Reasoner training was done using Apache Spark \cite{zaharia2016spark}. The Retriever and Reasoner were trained sequentially, each for 10 epochs on a single Tesla V100 GPU, with a total training time of approximately 1 week.

The final \acrshort{ours} Retriever architecture, as well as the MLM baseline, consisted of a scaled down version of PubMedBERT \cite{Gu_2021} trained from scratch on the task described in \ref{methods-mlm}, with 4 hidden layers, 4 attention heads, an intermediate size of 512, a hidden size of 256, and total size of 10 million parameters. Final Retriever/MLM training used a batch size of 512, a categorical cross-entropy loss, and an AdamW optimizer \cite{loshchilov2019decoupled} with a learning rate of 0.0001 and no weight decay.

The \acrshort{ours} architecture is summarised in Figure \ref{figure-easl-model}. The final query-evidence encoder component of the \acrshort{ours} Reasoner architecture consisted of a layer normalisation across all concatenated query/evidence pairs, then two conv1d layers each with kernel size of 1 (first layer: 2 input channels, 8 output channels; second layer: 8 input channels, 1 output channel) across each query/evidence pair individually. The final evidence combiner component of the \acrshort{ours} Reasoner architecture consisted of a set transformer \cite{lee2019set} over all query-evidence embeddings returning a single embedding, followed by a linear layer and sigmoid to output a binary probability. The set transformer had 4 heads, 2 induced set attention blocks with 32 inducing points for the encoder, and a pooling by multihead attention followed by two set attention blocks in the decoder. The Reasoner had a total size of 2 million parameters. After freezing the Retriever weights, the final Reasoner training used a batch size of 2048, binary cross-entropy loss, and AdamW optimizer with a learning rate of 0.0001 and weight decay of 0.001. For both training and inference, 64 evidence sentences were retrieved for a given query. A post-hoc frequency bias correction factor of 0.5 was used for the \acrshort{ours}-cor variant (Section \ref{methods-post-hoc-correction} and Appendix \ref{appendix-bias-corr-theory} for details of post-hoc correction).

\begin{table}[t!]
\caption{\textbf{\acrshort{ours} hyperparameter summary}}
\label{tab:hyperparameters}
\vskip 0.15in
\begin{center}
\begin{small}
\begin{sc}
\begin{tabular}{>{\raggedright\arraybackslash}p{5cm}|p{9cm}}
\toprule
\textbf{Component} & \textbf{Hyperparameters} \\
\midrule

\textsc{Tokenization} &
\begin{itemize}[leftmargin=*, noitemsep, topsep=0pt]
  \item Max sequence length: 128
  \item Truncated and padded
  \item Tokenized using PubMedBERT vocabulary
\end{itemize} \\

\hline
\textsc{Training Setup} &
\begin{itemize}[leftmargin=*, noitemsep, topsep=0pt]
  \item 10 epochs for Retriever
  \item 10 epochs for Reasoner
  \item Single Tesla V100 GPU
  \item Total training time: approx. 1 week
\end{itemize} \\

\hline
\textsc{Retriever / MLM Baseline} &
\begin{itemize}[leftmargin=*, noitemsep, topsep=0pt]
  \item \textbf{Architecture:}
  \begin{itemize}[leftmargin=1.5em]
    \item 4 hidden layers, 4 attention heads
    \item Hidden size: 256, Intermediate size: 512
    \item Total parameters: approx. 10M
  \end{itemize}
  \item \textbf{Training:}
  \begin{itemize}[leftmargin=1.5em]
    \item Batch size: 512 samples
    \item Loss: Categorical cross-entropy
    \item Optimizer: AdamW
    \item Learning rate: 0.0001, Weight decay: 0.0
  \end{itemize}
\end{itemize} \\

\hline
\textsc{Reasoner} &
\begin{itemize}[leftmargin=*, noitemsep, topsep=0pt]
  \item \textbf{Encoder:}
  \begin{itemize}[leftmargin=1.5em]
    \item LayerNorm over concatenated query-evidence pairs
    \item Two Conv1D layers:
    \begin{itemize}
      \item 1st: 2 $\rightarrow$ 8 channels
      \item 2nd: 8 $\rightarrow$ 1 channel
    \end{itemize}
  \end{itemize}
  \item \textbf{Combiner (Set Transformer):}
  \begin{itemize}[leftmargin=1.5em]
    \item 4 heads
    \item 2 ISABs (encoder) with 32 inducing points
    \item PMA pooling
    \item 2 SABs (decoder)
  \end{itemize}
  \item \textbf{Output:} Linear layer + sigmoid
  \item \textbf{Total parameters:} approx. 2M
  \item \textbf{Training:}
  \begin{itemize}[leftmargin=1.5em]
    \item Batch size: 2048 samples
    \item Loss: Binary cross-entropy
    \item Optimizer: AdamW
    \item Learning rate: 0.0001, Weight decay: 0.001
  \end{itemize}
\end{itemize} \\

\hline
\textsc{Retriever Inference} &
\begin{itemize}[leftmargin=*, noitemsep, topsep=0pt]
  \item 64 evidence sentences retrieved per query
\end{itemize} \\

\hline
\textsc{Bias Correction} &
\begin{itemize}[leftmargin=*, noitemsep, topsep=0pt]
  \item \textbf{Post-hoc frequency bias correction factor:}
  \begin{itemize}[leftmargin=1.5em]
    \item 0.5 for \acrshort{ours}-cor variant
    \item 0.0 for \acrshort{ours}-uncor variant
  \end{itemize}
\end{itemize} \\

\bottomrule
\end{tabular}
\end{sc}
\end{small}
\end{center}
\vskip -0.1in
\end{table}

The post-hoc frequency bias correction factor selection and architectural comparison ablations (Appendix \ref{appendix-ablations}) were based on MRR for a 2020 year split model, on a \textit{Held-out Biomedical Literature} validation set containing one biomedical literature cloze-style query sentence per gene in \textit{Genes} from publicly-available abstract sections (Appendix \ref{appendix-lit-dataset}). The resulting 15477 validation set queries were therefore sentences published from 2020 onwards, and retrieval corpus sentences published prior to 2020. The learning rate was chosen to reduce training time while maintaining training stability, and the batch size selected to optimise GPU utilisation. We did not evaluate variations of model scale and leave this to future work.

\section{Approximating Evidence Shapley Values}
\label{appendix-shap}

We used a simple Monte Carlo method to approximate Shapley values, combined with antithetical sampling for variance reduction~\cite{mitchell2022sampling}. The Shapley value was approximated as 
\begin{equation}
\phi_i \approx \frac{1}{2M} \sum_{j=1}^{M} \left( \left[g(S_j \cup \{i\}) - g(S_j)\right] + \left[g(\bar{S}_j \cup \{i\}) - g(\bar{S}_j)\right] \right)
\end{equation}
where $\phi_i$ is the approximate Shapley value of feature $i$ (an encoded query/evidence pair), $M$ is the chosen number of sampled permutations, $S_j$ is the set of features preceding $i$ in the j-$th$ permutation sample, $g(S_j)$ is the Reasoner output when only the features $S_j$ are unmasked, $g(S_j \cup \{i\})$ is the Reasoner output when feature $i$ is unmasked in addition to $S_j$, and $\bar{S}_j$ corresponds to the set of features preceding $i$ in the reverse of the j-$th$ permutation sample (equivalently, the set of features following $i$ in the j-$th$ permutation sample). The sum of the Shapley values over features plus the score when all features are $\mathrm{NULL}$ equates to the final score. Depending on the purpose, we use either the post-sigmoid output or the logit score for $g$. We use $M=100$ whenever Shapley values are computed as part of this paper. See Appendix \ref{appendix-speed} for profiling of Shapley computation.

\begin{algorithm}[H]
\caption{Generate permutation-approximated Shapley attributions for a single query.}\label{alg:shap}
\begin{algorithmic}
\State \textbf{Input:} Number of permutations $M$, query-evidence embeddings $ E = \{d^{qe}_1, \ldots, d^{qe}_k\} $, missing evidence embedding $NULL$, model forward function $g(\cdot)$
\State \textbf{Output:} Shapley value of query-evidence embeddings: $\phi_1, \ldots, \phi_k$
\State Initialize $\phi_i = 0$ for $i= 1,\ldots,k$
\State $2M$ antithetical sample of permutations $p^j$ for $j \in 1, \ldots , 2M$ of the feature indices $1,\ldots,k$
\State where $p^{M+i} = \mathrm{ReverseOrder}(p^i)$
\State $\tilde{E}_0 = \{\mathrm{NULL}, \ldots ,\mathrm{NULL}\}$, with $|\tilde{E}_0| = k$
\State $s_0 \leftarrow g(\tilde{E}_0)$
\For{$ j \in \{1, \ldots, 2M\} $}
  \For{$ i \in \{1, \ldots, k\} $}
    \State $\tilde{E}^j_i \leftarrow \{d^{qe}_{p^j[1]},\ldots,d^{qe}_{p^j[i]}, \mathrm{NULL}, \ldots, \mathrm{NULL}\}$, with $|\tilde{E}^j_i| = k$
    \State $s_i \leftarrow g(\tilde{E}^j_i)$
    \State $\phi_{p^j[i]} \leftarrow \frac{j-1}{j}\phi_{p^j[i]} + \frac{1}{j}(s_i - s_{i-1})$ \\ \Comment{Cumulative average of marginals for feature $p^j[i]$ across permutations}
  \EndFor
\EndFor
\end{algorithmic}
\end{algorithm}

\section{Relationship between Multinomial and Binary Objectives}
\label{appendix-objective-theory}
\acrshort{ours} is trained to predict the probability that a given query-entity pair is ``true", i.e. that it came from a real occurrence in the literature and was not randomly generated. Given the labels $L \in \{0, 1\}$, the query (masked sentence) variable $Q$, the named entity answer variable $A$, the Reasoner parameters $\theta$ and the fixed Retriever parameters $\psi$, the model is trained to predict
\begin{align}
    \frac{1}{1+\exp(-z(a_i,q_i))} \approx P(L=1 | Q=q_i, A=a_i; \theta, \psi)
\end{align}
where $z(a_i,q_i)$ is the logit output of the network in response to a specific example $i$, i.e.
\begin{align}
    z(a_i,q_i) \approx &\log(P(L=1 | Q=q_i, A=a_i; \theta, \psi)) \label{eq:obj} \\
          - &\log(P(L=0 | Q=q_i, A=a_i; \theta, \psi)) \notag.
\end{align}
Here, when $L=0$, the example $i$ corresponds to a negative example where $Q$ and $A$ have been chosen independently. Consider the case where the specific parameters $\theta$ and $\psi$ have been learned such that the equality in Eq.~\ref{eq:obj} holds exactly; we are interested in the output in this case. We therefore assume the optimal output $z^*(a_i,q_i)$ and exclude the parameters.

The equation can be re-written using Bayes' Theorem,
\begin{align}
    z^*(a_i,q_i) = &\log(P(Q=q_i, A=a_i | L=1)) + \log(p(L=1)) - \log(P(Q=q_i, A=a_i)) \label{eq:logit} \\
          - &\log(P(Q=q_i, A=a_i | L=0)) - \log(p(L=0)) + \log(P(Q=q_i, A=a_i)).\notag
\end{align}
In our training setup, positive and negative examples are sampled equally often, i.e. 
\begin{align}
    \log(p(L=1)) = \log(p(L=0)).
\end{align}
As a result, Eq.~\ref{eq:logit} simplifies to 
\begin{align}
    z^*(a_i,q_i) = &\log(P(Q=q_i, A=a_i | L=1)) - \log(P(Q=q_i, A=a_i | L=0)) \label{eq:logit_simp} \\
\end{align}
Using the product rule
\begin{align}
    z^*(a_i,q_i) = &\log(P(A=a_i | Q=q_i, L=1)) + \log(P(Q=q_i | L=1))\notag \\
        - &\log(P(A=a_i | Q=q_i, L=0)) - \log(P(Q=q_i | L=0)).\notag 
\end{align}
The distribution over queries is also equal for positive and negative labels, as each query sentence is chosen for each condition once per epoch, simplifying to 
\begin{align}
    z^*(a_i,q_i) = &\log(P(A=a_i | Q=q_i, L=1)) - \log(P(A=a_i | Q=q_i, L=0)). 
\end{align}
The distribution over named entity answers is independent of the query when conditioned on $L=0$, because negative samples are chosen by randomly pairing queries and entities. So the second term here corresponds to our negative sampling distribution. Therefore, the output at optimality corresponds to 
\begin{align}
    z^*(a_i,q_i) &= \log(P(A=a_i | Q=q_i, L=1)) - \log(P(A=a_i | L=0)) \label{eq:ent_obj0} \\
     &= \log(P(A=a_i | Q=q_i, L=1)) + \log (|\mathcal{A}|) \notag
\end{align}
since the probability of choosing a given answer $a_i$ as a negative sample during training is $\frac{1}{|\mathcal{A}|}$. Comparing to the optimal logit output of the MLM model, we see a close relationship:
\begin{align}
    z^{*,\mathrm{MLM}}(a_i,q_i) = &\log(P(A=a_i | Q=q_i, L=1)) + \log(Z) \label{eq:ent_mlm_obj}
\end{align}
where $Z$ is the partition function (the MLM includes $L=1$ implicitly as all examples are positive). The optimal logit outputs for the models therefore scale up to their respective normalization factors.

\section{Post-hoc Frequency Bias Correction as Trading off Log Probability and Mutual Information}
\label{appendix-bias-corr-theory}

From Equation \ref{eq:ent_obj0} in Appendix \ref{appendix-objective-theory}, we find that the optimal model output logit scales with $\log(P(A | Q, L=1))$, i.e. the probability of the answer given the query assuming a real example ($L=1$). This score will be highly correlated with the prior distribution over the answer set, particularly for an imbalanced dataset (like the mentions of \textit{Genes} in the scientific literature corpus used in the paper).

One approach to counteract the literature bias, if desired, is to instead consider the pointwise mutual information between a given answer and a given query:
\begin{align}
    \mathrm{PMI}(A=a;Q=q) = \log{\frac{P(A=a|Q=q)}{P(A=a)}} \label{eq:corr_pmi}.
\end{align}
PMI is widely used in the NLP community to measure associations between keywords in a corpus, based on their marginal occurrence counts and joint co-occurrence counts \cite{jurafsky2019speech}. Similarly, we find that it offers a straightforward means of correcting for class imbalance after training the model.

For a model that predicts a multiclass output (like the MLM), we can directly adapt the output. Specifically, after normalizing the outputs to remove $\log(Z)$, where Z is the partition function,
\begin{align}
    z^{\mathrm{MLM}}_c(a_i,q_i) &= z^{\mathrm{MLM}}(a_i,q_i) - c \cdot \log P(A=a_i|L=1) \\ 
    &\approx \log{P(A=a_i|Q=q_i,L=1)} - c \cdot \log P(A=a_i|L=1) \notag
\end{align}

where $P(A=a_i|L=1)$ is estimated by the proportion of passages in the corpus where $a_i$ is the correct answer. When $c=0.0$, the two approaches are equivalent; while when $c=1.0$, the output approximates the PMI score in Equation \ref{eq:corr_pmi}. Stronger corrections penalize common answers, and the score is only positive if the model's estimated answer probability for the given query is higher than the frequency-based prior.

In \acrshort{ours}, we instead note that the optimal logit score in Equation \ref{eq:ent_obj0} already reflects PMI \emph{if} the negative sampling probability $P(A=a_i|L_i=0)$ was chosen to reflect the prior distribution over answers in the dataset, $P(A=a_i|L=1)$. We therefore consider a negative distribution $P_c(A=a_i|L=0)$ that trades off between a uniform distribution $\frac{1}{|\mathcal{M}|}$ and one based on the answer prior in the training corpus:
\begin{align}
    P_c(A=a_i|L=0) = \frac{C(a_i)^c}{\sum_{i=1}^{|\mathcal{A}|} C(a_i)^c} \label{eq:corr_neg}
\end{align}
where $C(a_i)$ is the count of occurrences of answer $a_i$ as a masked entity in the training corpus. When $c=1$, this corresponds to the background distribution of $a_i$ in the training corpus $P(A=a_i|L=1)$; when $c=0$, it corresponds to the uniform distribution $\frac{1}{|\mathcal{A}|}$.

One possible approach to bias correction is to set a fixed $c$ during training and use the resulting negative sampling distribution in Equation \ref{eq:corr_neg}. However, this approach grants less flexibility in terms of the desired bias correction at inference time. We therefore continue to use the fixed uniform distribution $\frac{1}{|\mathcal{A}|}$ and instead introduce a correction factor 
\begin{align}
    f_c = \log{\frac{1}{|\mathcal{A}|}} - \log{P_c(A=a_i|L=0)}.
\end{align}
Applying this correction to the logit output of \acrshort{ours} after training (Equation \ref{eq:ent_obj0}) yields 
\begin{align}
    z(a_i,q_i) + f_c \approx \log(P(A=a_i | Q=q_i, L=1)) - \log{P_c(A=a_i|L=0)}
\end{align}
which reflects a log probability estimate when $c=0$ and a pointwise mutual information estimate when $c=1$. We found that the best performance in terms of MRR on the \textit{Held-out Biomedical Literature} validation dataset (Appendix \ref{appendix-lit-dataset}), was achieved with a partial correction of $c=0.5$. We refer to this as \acrshort{ours}-cor, and refer to the case with $c=0.0$ as \acrshort{ours}-uncor.

The bias correction can be straightforwardly identified as an additional additive feature during Shapley value estimation to communicate its impact to the user. For under-represented answers, it can be seen as compensating for ``missing'' evidence, e.g. due to the lack of research on a particular target.

\section{\acrshort{ours} Inference Speed}
\label{appendix-speed}

We profiled \acrshort{ours} for both prediction and explanation. We used CPUs only, though GPUs could be used to achieve additional speed-ups by reducing the time taken for the forward pass.

\subsection{Prediction}
For prediction on CPUs, the MLM baseline took $\sim$140ms over one query on one core, obtaining scores for all 19,176 genes via a single forward pass. By comparison, the non-negligible components of \acrshort{ours} inference time are:
\begin{enumerate}
    \item The batched forward pass over 19,176 query-evidence pairs (one for each gene), through the Reasoner - $\sim$7.4s on one core, and scales linearly with cores
    \item Vector searches over the 19,176 FAISS indices corresponding to each gene, for the Retriever - $\sim$27s on one core, ~1.5s on 40 cores or \textless0.15s if one core per index
\end{enumerate}

 Since the evidence is split into separate retrieval indices for each of the potential answers, the top evidence from each can be found in parallel. Therefore, search can generally scale more efficiently than for a traditional single FAISS index. To optimise inference, the forward pass should be run in batches while the search results for each potential answer are returned from each corresponding FAISS index. As a result, the total time is then largely defined by the maximum time for the above two stages of batched forward pass and vector search, given the relevant parallelisation.

These results assume exact brute force vector search (IndexFlatIP search indices from FAISS \cite{johnson2019billion}) with a complexity of $O(nd)$, where $n$ is the number of vectors in the given search index and $d$ is the dimensionality of each vector. While vector search was not a bottleneck in our setup, if inference speed were a concern as the retrieval corpus scales, there are many out-of-the-box options for more efficient approximate nearest neighbour search indices, including within FAISS. The \acrshort{ours} profiling results above also assume access to a machine with $\sim$300GB memory for the FAISS indices; fast inference is achieved on widely available resources.

\subsection{Explanation}
For inference time explanations, we compute Shapley values using the permutation-based method detailed in Appendix \ref{appendix-shap}, using $M=100$ permutations (200 with antithetical sampling). With 64 evidence sentences retrieved for a given query, this results in 12,800 evidence set variations required to compute all 64 Shapley values. Therefore, \textless10 forward passes are required, with a reasonable batch size. Given the small size of the Reasoner module (2 million parameters), generating an explanation takes $\sim$5 seconds using a single CPU only.

We also note that more efficient methods exist for approximating Shapley values \cite{lundberg2017unified}, particularly for deep networks. However, since Shapley value efficiency is neither our primary focus nor prohibitive, we used a permutation-based approach (Appendix \ref{appendix-shap}).

\section{Evaluation Dataset Statistics}
\label{appendix-eval-data-sizes}
The total sizes of all test/evaluation datasets are shown in Table \ref{dataset-statistics-table}.

\begin{table}[!ht]
\caption{\textbf{Evaluation dataset statistics}}
\label{dataset-statistics-table}
\vskip 0.25in
\begin{center}
\begin{small}
\begin{sc}
\begin{tabular}{>{\raggedright\arraybackslash}p{4cm}c|c} 
\toprule
Dataset & Subset & Count \\
\midrule
\multirow{3}{4cm}{Held-out Biomedical Literature} & 2020 & 14429 \\
                                                  & 2021 & 14859 \\
                                                  & 2022 & 15074 \\
\hline
Gene Description Facts                                 &  & 60839 \\
\hline
\multirow{3}{4cm}{GDF Evidence Annotations (human expert)}     & Query-gene pairs & 8 \\
                                                  & Positives:Negatives & 4:4 \\
                                                  & Evidence & 512 \\
\hline
\multirow{3}{4cm}{GDF Evidence Annotations (GPT-4)}     & Query-gene pairs & 100 \\
                                                  & Positives:Negatives & 50:50 \\
                                                  & Evidence & 6400 \\
\hline
\multirow{2}{4cm}{Clinical Trial Outcomes (2005 onwards)}        & Successes & 1449 \\
                                                  & Fails & 4222 \\
\hline
\multirow{3}{4cm}{CTO Evidence Annotations (human expert)}     & Query-target pairs & 8 \\
                                                  & Successes:Fails & 4:4 \\
                                                  & Evidence & 512 \\
\hline
\multirow{3}{4cm}{CTO Evidence Annotations (GPT-4)}     & Query-target pairs & 100 \\
                                                  & Successes:Fails & 50:50 \\
                                                  & Evidence & 6400 \\
\bottomrule
\end{tabular}
\end{sc}
\end{small}
\end{center}
\vskip -0.1in
\end{table}

\section{Predicting Genes in Held-out Biomedical Literature}
\label{appendix-bml}

\paragraph{Dataset Construction} For all experiments in this section, we trained the MLM (\acrshort{ours} Retriever) and \acrshort{ours} Reasoner only on biomedical literature data published prior to 2020. Except where specified, \acrshort{ours} also only retrieved data published prior to 2020 (Figure \ref{figure-data-splits}). We then constructed a \textit{Held-out Biomedical Literature} evaluation dataset from publicly-available paper abstracts. We generated a balanced dataset to obtain results reflecting performance across all 19,176 genes, not biased to the most well-studied (discussed further in Appendix \ref{appendix-non-strat}). We sampled one sentence per unique gene in \textit{Genes} for each of the years 2020, 2021, and 2022; further details in Appendix \ref{appendix-lit-dataset}.

\paragraph{Results} \acrshort{ours} improved on the baselines over all year subsets, both with and without bias correction (Table \ref{literature-table}). Bias-corrected \acrshort{ours} improved on uncorrected performance, consistent with the use of a balanced evaluation dataset. For completeness, we show results on an imbalanced dataset (without stratification by gene in \textit{Genes}) in Appendix \ref{appendix-non-strat}.

To test \acrshort{ours}'s ability to leverage retrieved literature that it was not trained on, we enabled retrieval up to the year preceding the query sentence publication (rather than strictly prior to the 2020 training data cutoff). This improved performance (\acrshort{ours}-cor-updated, Table \ref{literature-table}).

\begin{table*}[ht!]
\caption{\textbf{Held-out Biomedical Literature}: Ranking metrics on a dataset consisting of one sentence per gene in \textit{Genes} for each year of 2020, 2021, and 2022. MLM and \acrshort{ours} trained on data published prior to 2020. MCS, \acrshort{ours}-uncor and \acrshort{ours}-cor also retrieved data published prior to 2020. \acrshort{ours}-cor-updated retrieved up to the year before the publication year of the query sentence.}
\label{literature-table}
\vskip 0.15in
\begin{center}
\begin{small}
\begin{sc}
\begin{tabular}{lc|ccc|cc||c}
\toprule
Metric & Query year & \multicolumn{3}{c|}{Baselines} & \multicolumn{3}{c}{\acrshort{ours}}\\
       &      & Freq & MCS & MLM & Uncor & Cor & Cor-Updated \\
\midrule
       & 2020 & $<$0.001 & 0.182 & 0.181 & 0.198 & \textbf{0.233} & - \\
MRR    & 2021 & $<$0.001 & 0.172 & 0.169 & 0.187 & \textbf{0.215} & \textbf{0.223} \\
       & 2022 & $<$0.001 & 0.167 & 0.164 & 0.178 & \textbf{0.205} & \textbf{0.219} \\
\hline
       & 2020 & 7661 & 3280 & 3465 & 2803 & \textbf{2489} & - \\
MR     & 2021 & 7834 & 3568 & 3789 & 3032 & \textbf{2695} & \textbf{2544} \\
       & 2022 & 7931 & 3770 & 4016 & 3287 & \textbf{2902} & \textbf{2623} \\
\hline
        & 2020 & $<$0.001 & 0.268 & 0.269 & 0.291 & \textbf{0.333} & - \\
h@10 & 2021 & $<$0.001 & 0.251 & 0.252 & 0.274 & \textbf{0.313} & \textbf{0.324} \\
        & 2022 & $<$0.001 & 0.243 & 0.243 & 0.260 & \textbf{0.295} & \textbf{0.312} \\
\hline
         & 2020 & 0.014 & 0.443 & 0.438 & 0.484 & \textbf{0.521} & - \\
h@200 & 2021 & 0.014 & 0.422 & 0.416 & 0.456 & \textbf{0.497} & \textbf{0.509} \\
         & 2022 & 0.013 & 0.404 & 0.398 & 0.435 & \textbf{0.473} & \textbf{0.496} \\
\bottomrule
\end{tabular}
\end{sc}
\end{small}
\end{center}
\vskip -0.1in
\end{table*}

\section{Comparison of Models on a Non-Stratified Held-out Biomedical Literature Dataset}
\label{appendix-non-strat}

Gene mention counts are extremely imbalanced in the literature. In the training data, of the 19,176 protein-coding genes, the most-well studied has approximately 2 million mentions, while the least studied 10,000 genes all have less than 1,000 mentions. For our \textit{Held-out Biomedical Literature} dataset we used stratified sampling (stratification by gene in \textit{Genes}) to obtain a class balanced test dataset, with equal counts of each gene to avoid dominance of well-studied genes. By preventing reliance of models on the gene frequency distribution prior, a class-balanced setup is especially challenging. Strong performance across the genome is desirable because understudied genes are of particular interest in drug discovery, when seeking new ways to treat a disease. 

While our focus is therefore on balanced performance across the genome (results in Appendix \ref{appendix-bml}), for completeness, we also evaluated \acrshort{ours} on a dataset of 20,000 randomly-sampled publicly-available abstract sentences published from 2020 onwards, obtaining an imbalanced dataset \textit{without} stratification by gene in \textit{Genes}. As expected, the frequency-based baseline performs significantly better here relative to the stratified dataset in Table \ref{literature-table}, reflecting that ability to rely on the frequency distribution prior. Ranking metrics show similar performance for \acrshort{ours}, MCS and MLM (Table \ref{not-strat-table}). In comparison, on the more challenging stratified setup \acrshort{ours} markedly outperforms baselines (Table \ref{literature-table}). Comparing \acrshort{ours} and MLM, \acrshort{ours}'s superior balanced performance across the genome could be explained by it's access to a knowledge base even for the most rare genes, avoiding the need to memorise knowledge of genes rarely seen at training time in the model parameters. \acrshort{ours} obtains superior performance on less studied genes without sacrificing performance on well-studied genes.

\begin{table*}[ht!]
\caption{\textbf{Non-stratified Held-out Biomedical literature}: \acrshort{ours} ranking metrics on a random subsplit (not stratified by gene in \textit{Genes}) of query sentences published from 2020 onwards (20,000 queries), for an \acrshort{ours} model trained and retrieving from data prior to 2020.}
\label{not-strat-table}
\vskip 0.15in
\begin{center}
\begin{small}
\begin{sc}
\begin{tabular}{l|ccc|cc}
\toprule
Metric & \multicolumn{3}{c|}{Baselines} & \multicolumn{2}{c}{\acrshort{ours}} \\
       & Freq & MCS & MLM & Uncor & Cor \\
\midrule
MRR    & 0.026 & \textbf{0.405} & 0.399 & 0.403 & 0.350 \\
MR     & 2321 & \textbf{1114} & 1305 & 1140 & 1456\\
h@10 & 0.070 & 0.520 & 0.519 & \textbf{0.523} & 0.500 \\
h@200 & 0.304 & 0.691 & 0.686 & \textbf{0.699} & 0.686 \\
\bottomrule
\end{tabular}
\end{sc}
\end{small}
\end{center}
\vskip -0.1in
\end{table*}

\section{Architecture Ablation Experiments}
\label{appendix-ablations}

We performed ablations of all core \acrshort{ours} architectural components, including the Reasoner, Retriever and frequency bias correction. A \textit{Held-out Biomedical Literature} validation set was used for ablations experiments, consisting of one sentence per gene in \textit{Genes} sampled from publicly-available abstract sentences published from 2020 onwards (as described in Appendices \ref{appendix-lit-dataset} \& \ref{appendix-easl-hyperparameters}), for an \acrshort{ours} model trained and retrieving from data prior to 2020. The results are summarised in Table \ref{ablations-table}. All ablations resulted in a drop in performance across all ranking metrics, demonstrating the benefit of \acrshort{ours} components.

\begin{table*}[ht!]
\caption{\textbf{Architecture ablations}: Ablated versions of \acrshort{ours}-uncor on a validation dataset consisting of one sentence per gene in \textit{Genes} sampled from sentences published from 2020 onwards, while training on and retrieving from data prior to 2020. Hadamard: substituting the convolution layers of the Reasoner with a Hadamard product. PubMedBERT: substituting the Retriever for the PubMedBERT model.}
\label{ablations-table}
\vskip 0.15in
\begin{center}
\begin{small}
\begin{sc}
\begin{tabular}{l|cc|cc|cc}
\toprule
Metric & \multicolumn{2}{c|}{\acrshort{ours}} & \multicolumn{2}{c|}{Reasoner Ablations} & \multicolumn{2}{c}{Retrieval Ablations} \\
       & Cor & Uncor & MCS & Hadamard & PubMedBERT & MLM \\
\midrule
MRR    & \textbf{0.211} & 0.181 & 0.163 & 0.166 & 0.134 & 0.163 \\
MR     & \textbf{2873} & 3210 & 3726 & 3260 & 3606 & 3945 \\
h@10 & \textbf{0.302} & 0.262 & 0.241 & 0.253 & 0.207 & 0.242 \\
h@200 & \textbf{0.482} & 0.443 & 0.409 & 0.441 & 0.389 & 0.404 \\
\bottomrule
\end{tabular}
\end{sc}
\end{small}
\end{center}
\vskip -0.1in
\end{table*}

\subsection{Reasoner}

The MCS baseline (Section \ref{results-baselines-ablations}) acts as an ablation of the \acrshort{ours} Reasoner, since it relies solely on query-evidence cosine similarities of the Retriever to obtain a score.

We also selectively ablated the convolutional query-evidence encoder component of the \acrshort{ours} Reasoner (Section \ref{methods-reasoner}) by substituting that component for a parameter-free Hadamard product between the query embedding and each evidence embedding. The Hadamard product was chosen in order to incorporate an inductive bias towards the cosine similarity.

\subsection{Retriever}

We ablated our task specific Retriever (Sections \ref{methods-mlm} \& \ref{methods-retriever-architecture}), by replacing it with an off-the-shelf biomedical transformer. Specifically we used a PubMedBERT model initialised with its published weights \cite{Gu_2021} as the Retriever. We created sentence embeddings by taking the mean over outputs corresponding to \lbrack MASK\rbrack{} tokens. This Retriever had a larger hidden size with 768 dimensional query and evidence embeddings. The \acrshort{ours} Reasoner was therefore linearly scaled to match this hidden size.

We also evaluated the MLM baseline (Section \ref{methods-mlm}), which acts as an ablation of R2E in its entirety, taking a fully parametric approach to prediction.

\subsection{Post-hoc frequency bias correction}

We report results with and without bias correction.

\section{Further Details on Creation of Gene Description Facts Dataset}
\label{appendix-gene-descriptions-creation}

We downloaded UniProt FTP server data version 2023\_01 and extracted descriptions of protein functions for our set of protein-coding gene entities (\textit{Genes}) from \href{https://www.uniprot.org/}{UniProt (Universal Protein Resource)}, used under \href{https://creativecommons.org/licenses/by/4.0/legalcode}{CC BY (4.0)},  \cite{uniprot2022} (by pulling ``text'' from UniProt entities with type ``function'' in the ``comment'' field). Each entry is a human-written description of function, and consists of one or more sentences.

After dropping all descriptions containing fewer than four words, we converted each description into a set of single-sentence facts as follows:

\begin{enumerate}
    \item Descriptions were split into individual sentences and PubMed IDs removed, using regular expression operations.
    \item Each sentence was converted into a fact containing a ``\lbrack MASK\rbrack{}'' referring to the gene and ``\lbrack MASK\rbrack{}'' in place of all gene mentions, using one-shot prompted Claude 2.0 language model from Anthropic (prompt template below). Sentences which Claude determined did not contain a suitable fact, were dropped.
    \item ``\lbrack MASK\rbrack{}''-containing facts were extracted from the Claude completion, and facts without any ``\lbrack MASK\rbrack{}'' mention were dropped.
\end{enumerate}

For example, the description for the protein corresponding to gene \textit{ELF2} is:

\begin{quotation}
\noindent``Isoform 1 transcriptionally activates the LYN and BLK promoters and acts synergistically with RUNX1 to transactivate the BLK promoter. Isoform 2 may function in repression of RUNX1-mediated transactivation.''
\end{quotation}

From this description, the following facts were extracted for the evaluation dataset:

\begin{itemize}
    \item \lbrack MASK\rbrack{} isoform 1 transcriptionally activates the LYN and BLK promoters and acts synergistically with RUNX1 to transactivate the BLK promoter.
    \item \lbrack MASK\rbrack{} isoform 2 may function in repression of RUNX1-mediated transactivation.
\end{itemize}

The following one-shot prompt template was used to convert sentences from pulled UniProt gene descriptions into \lbrack MASK\rbrack{}-containing facts. The gene GENE\_NAME and UNIPROT\_DESCRIPTION\_SENTENCES were substituted into the template for each sentence-gene pair in the dataset, prior to querying Claude 2.0 via Anthropic's API.

\begin{verbatim}
    {HUMAN_PROMPT}
    # THE TASK:
    You are an expert biologist. You will be given a set of sentences from a 
    DESCRIPTION of a GENE from UniProt.
    
    Your instructions are to go one-by-one through each sentence in the 
    DESCRIPTION, and:
    1. If the sentence states a fact about the specified GENE convert the 
    sentence into a FACT according to the FACT formatting shown in the <example> 
    below. 2. If, and only if, the sentence does not state any information 
    about the GENE, you may skip the sentence and indicate this with 
    "sentence[nb] SKIPPED" as shown in the <example> below.
    
    # FORMATTING:
    Here's an example input and output contained in the <example> XML tags, 
    to illustrate the format in which FACTs should be stated, including how to 
    indicate that a sentence has been skipped.
    
    <example>
    Input:
    GENE: PGP
    DESCRIPTION sentences: 
    <sentence1>Glycerol-3-phosphate phosphatase hydrolyzing glycerol-3-phosphate 
    into glycerol.</sentence1>
    <sentence2>Thereby, regulates the cellular levels of glycerol-3-phosphate a 
    metabolic intermediate of glucose, lipid and energy metabolism.<\sentence2>
    <sentence3>Was also shown to have a 2-phosphoglycolate phosphatase activity 
    and a tyrosine-protein phosphatase activity.</sentence3> 
    <sentence4>However, their physiological relevance is unclear 
    (PubMed:26755581).</sentence4>
    <sentence5>In vitro, has also a phosphatase activity toward ADP, ATP, GDP 
    and GTP (By similarity).</sentence5>
    <sentence6>Further work is needed to understand this.</sentence6>
    <sentence7>(Microbial infection) Involved in replication of Rubella virus.
    </sentence7>
    
    Output:
    Here are complete set of [MASK]-containing FACTs for each sentence about PGP:
    <sentence1_fact>[MASK] is a glycerol-3-phosphate phosphatase that hydrolyzes 
    glycerol-3-phosphate into glycerol.</sentence1_fact>
    <sentence2_fact>[MASK] regulates cellular levels of glycerol-3-phosphate, a 
    metabolic intermediate of glucose, lipid and energy metabolism.
    </sentence2_fact>
    <sentence3_fact>[MASK] has 2-phosphoglycolate phosphatase activity and 
    tyrosine-protein phosphatase activity.</sentence3_fact>
    <sentence4_fact>sentence4 SKIPPED</sentence4_fact>
    <sentence5_fact>In vitro, [MASK] has phosphatase activity toward ADP, ATP, 
    GDP and GTP.</sentence5_fact>
    <sentence6_fact>sentence6 SKIPPED</sentence6_fact>
    <sentence7_fact>[MASK] is involved in replication of Rubella virus.
    </sentence7_fact>
    </example>
    
    # FACT REQUIREMENTS
    You must note the following requirements, when constructing each FACT:
    1. Each and every FACT must include one or more [MASK] tokens representing 
    the GENE.
    2. All references to or synonyms of the GENE anywhere in a FACT, must also 
    be replaced by [MASK].
    3. Only include information explicitly stated in the DESCRIPTION sentence 
    when extracting a FACT - do not elaborate with any additional information 
    from elsewhere.
    4. You must go through every sentence.
    5. You can only skip a sentence if it contains no information about the 
    GENE, and you must indicate this by stating the sentence was SKIPPED in 
    the corresponding sentence FACT XML tags.
    
    # THE FINAL GENE AND DESCRIPTION SENTENCES
    Now, paying attention to all the above instructions and example, please go 
    one-by-one through each sentence in the following DESCRIPTION and extract 
    each FACT for the stated GENE:
    
    Input:
    GENE: {GENE_NAME}
    DESCRIPTION sentences:
    {UNIPROT_DESCRIPTION_SENTENCES}
    
    {AI_PROMPT}
    Output:
    Here are complete set of [MASK]-containing FACT(s) for each sentence about 
    {GENE_NAME}:
    <sentence1_fact>
\end{verbatim}

\section{Further Details on Creation of Explanation Annotations for Gene Description Facts Dataset}
\label{appendix-gdf-evidence-annotations}

We constructed \textit{Evidence Annotations} for the \textit{Gene Description Facts} dataset by having GPT-4 annotate query relevance for all evidence across 50 randomly sampled query-entity pairs (positive examples) and the same 50 queries but with a randomly sampled alternative entity (negative examples), resulting in 6400 query-evidence pairs (100 queries each retrieving 64 pieces of evidence) with a binary annotation. The same instructions were followed by the drug discovery expert when providing annotations used to validate the GPT-4 annotations in Section \ref{sec:gdf}. The expert annotator was a Principal Scientist with over two years industry drug target identification experience in addition to holding a domain-relevant PhD and post-doc. They consented to the use of their annotations.

We used GPT-4 to obtain relevant/irrelevant annotations for this task by using the following prompt, substituting in GENE\_DESCRIPTION\_FACT and EVIDENCE\_SENTENCE:

\begin{verbatim}
    You are a scientific expert working on target identification in drug 
    discovery.
    
    Your task is to use your expertise to evaluate whether a piece of evidence 
    (referred to as EVIDENCE) about a masked target from an academic paper (in 
    the form of a sentence), provides relevant support to a specified biological 
    fact about that masked target (referred to as FACT). Please explain your 
    reasoning first before giving your answer. Provide your final answer by 
    stating either <answer>RELEVANT</answer> or <answer>IRRELEVANT</answer> 
    to indicate whether the evidence sentence provides relevant support for 
    the FACT or not. Please also indicate your confidence in your answer by 
    writing one of <confidence>HIGH</confidence> or <confidence>LOW</confidence>.
    
    Here are some examples demonstrating proper formatting and reasoning in a 
    response:
    <example>
    H: Your FACT of interest is:
    <fact>
    [MASK] also inhibits Wnt signaling by binding to the CTNNB1 protein, 
    preventing interaction of CTNNB1 with TCF7L2/TCF4.
    </fact>
    
    Here is the EVIDENCE sentence to assess:
    <evidence>
    In the HTB-72 melanoma cell line, [MASK] inhibited melanoma cell growth 
    and additionally, [MASK] also induced apoptosis.
    </evidence>
    
    A:The evidence sentence makes no indication that [MASK] is involved in Wnt 
    signaling or anything else mentioned in the given biological fact. Therefore, 
    it is <answer>IRRELEVANT</answer> with <confidence>HIGH</confidence>.
    </example>
    
    <example>
    H: Your FACT of interest is:
    <fact>
    [MASK] is a probable serine protease that plays a role in hearing.
    </fact>
    
    Here is the EVIDENCE sentence to assess:
    <evidence>
    type hearing loss, as occurred in subject SB114-206, caught our attention 
    (Figure 1a) because alterations in [MASK], a serine protease, are also known 
    to cause progressive ski slope–type hearing loss.
    </evidence>
    
    A:The evidence sentence states that [MASK] is a serine protease that can affect 
    hearing so it directly supports our biological fact being true of [MASK]. 
    Therefore, it is <answer>RELEVANT</answer> with <confidence>HIGH</confidence>.
    </example>
    
    Now here is your real task. 
    Your FACT of interest is:
    <fact>
    {GENE_DESCRIPTION_FACT}
    </fact>
    
    Here is the EVIDENCE sentence to assess:
    <evidence>
    {EVIDENCE_SENTENCE}
    </evidence>
\end{verbatim}

\section{Examples of Evidence Annotations for Gene Description Facts Dataset}
\label{appendix-gdf-evidence-annotations-examples}

Here we show an example of each of true positive, true negative, false positive and false negative \acrshort{ours} evidence Shapley values versus GPT-4 relevant/irrelevant \textit{Evidence Annotations}, for \textit{Gene Description Facts}. These examples also illustrate the level of biological complexity in the \textit{Gene Description Facts} task.

\begin{enumerate}
\item \textbf{True positive}: \acrshort{ours} assigns a large, positive Shapley value to a piece of evidence that GPT-4 labels as relevant. \acrshort{ours} and GPT-4 recognise this as a relevant piece of evidence.

\begin{verbatim}
    Query: [MASK] regulates actin dynamics in motor neurons by inhibiting
    the GTP-binding activity of ARF6, leading to ARF6 inactivation.

    Target: C9orf72
    
    Evidence: The observation that the GTP-bound forms of Rac1 and Arf6 are
    increased when [MASK] is repressed in primary mouse motor neurons or
    in patient iPSC-derived motor neurons strongly suggests that [MASK] is
    essential for the GTP exchange activity of Arf6 in particular. 
    (PMID: 27723745)

    R2E-generated Shapley value: 0.0391

    GPT4 completion: A: The evidence sentence directly supports the biological
    fact. It states that when [MASK] is repressed, the GTP-bound forms of ARF6
    are increased, suggesting that [MASK] is essential for the GTP exchange 
    activity of ARF6. This aligns with the fact that [MASK] regulates actin 
    dynamics in motor neurons by inhibiting the GTP-binding activity of ARF6, 
    leading to ARF6 inactivation. Therefore, it is <answer>RELEVANT</answer> 
    with <confidence>HIGH</confidence>.
\end{verbatim}

\item \textbf{True negative}: \acrshort{ours} assigns a negative Shapley value to a piece of evidence that GPT-4 labels as irrelevant. \acrshort{ours} and GPT-4 recognise this as an irrelevant piece of evidence.

\begin{verbatim}
    Query: [MASK] binds the 3'-UTR of TRPV1 mRNA and represses TRPV1 
    translation, which is required to maintain normal thermoception.

    Target: DCTN6
    
    Evidence: Gene co-expression analyses for SLC2A4 were conducted with
    West Breast [ 28 ], and the results showed that SLC2A4 is co-expressed with
    FADD, BLOC151, RHOB, [MASK], CELF2, SNTB2, NPPB, TIE1, FGFR1, IDH1, ECH1 ,
    etc. (Fig. 5 D). (PMID: 34525987)

    R2E-generated Shapley value: -0.0074

    GPT4 completion: A: The evidence sentence only mentions that [MASK] is
    co-expressed with SLC2A4 and other genes, but it does not provide any
    information about [MASK] binding to the 3'-UTR of TRPV1 mRNA or
    repressing TRPV1 translation. Therefore, it is <answer>IRRELEVANT</answer>
    with <confidence>HIGH</confidence>.
\end{verbatim}

\item \textbf{False positive}: \acrshort{ours} assigns a large, positive Shapley value to a piece of evidence that GPT-4 labels as irrelevant. \acrshort{ours} seems to consider the mention of magnesium concentration in relation to [MASK] sufficient to support the fact that "[MASK] can use Mg (2+)".

\begin{verbatim}
    Query: [MASK] can use either Ca(2+) or Mg(2+) equally.

    Target: AGTR2

    Evidence: Magnesium concentrations were 2 mol/L, except for [MASK]
    (2.5 mol/L). (PMID: 11841622)

    R2E-generated Shapley value: 0.0981

    GPT4 completion: A: The evidence sentence only provides information
    about the magnesium concentration in relation to [MASK], but it does not
    provide any information about [MASK]'s ability to use either Ca(2+) or
    Mg(2+) equally. Therefore, it is <answer>IRRELEVANT</answer> with
    <confidence>HIGH</confidence>.
\end{verbatim}

\item \textbf{False negative}: \acrshort{ours} assigns a negative Shapley value to a piece of evidence that GPT-4 labels as relevant. \acrshort{ours} fails to recognize that this is a relevant piece of evidence.

\begin{verbatim}
    Query: [MASK] is required for mitochondrial tubular network
    and cristae organization.

    Target: GHITM

    Evidence: Decreased [MASK] expression and loss of cristae organisation 
    occur at an early stage of DIO and represent an IRF5-dependent mechanism
    that may contribute to loss of microenvironmental homeostasis and 
    development of insulin resistance (Fig. 8 ). (PMID: 36042203)

    R2E-generated Shapley value: -0.0090

    GPT4 completion: A: The evidence sentence states that decreased [MASK]
    expression leads to loss of cristae organization, which directly 
    supports the biological fact that [MASK] is required for cristae 
    organization. Therefore, it is <answer>RELEVANT</answer> with 
    <confidence>HIGH</confidence>.
\end{verbatim}
\end{enumerate}

\section{Further Details on Creation of Clinical Trial Outcomes Dataset}
\label{appendix-cod-creation}

Following the protocol described in \citealt{nelson2015support} and \citealt{minikel2023refining}, we filtered drugs from the commercial PharmaProjects \cite{pharmaprojects} drug development database. We filtered out drugs that were combination treatments and diagnostic drugs. In addition to the \citealt{nelson2015support} filtering protocol, we applied a further temporal filtering of drugs to avoid temporal leakage (Section \ref{sec:cod}). Each drug was attributed an 'earliest evidence year', the earliest year that could be extracted from a mix of free-text and structured data fields in each PharmaProjects drug record. All dates were extracted from either: a ``key events'' field, which has well structured but heterogeneously populated dates; or free text fields giving details about preclinical, Phase I, Phase II and Phase III development or a general description of a drug's development trajectory. From the free text fields, all 4 digit date-like strings which did not occur in contexts with common failure modes were extracted using the regex \verb|(?<=[^0-9a-zA-Z\=\%])([0-9]{4})(?=[\,\\\s\;)])(?![\s*m+g+l+])|. In brief, 4 digits, in brackets, followed by a comma, whitespace or backslash, and not subsequently followed by characters indicating quantitative measurements (namely `m', `g' and `l'). Anomalous dates introduced by the regex were removed by dropping any dates that were more than 50 years from the median of the dates for a drug record. Across all of these date fields the earliest date was attributed to the drug and all indications it was tested against and used to include or exclude drugs from the analysis. The earliest development date for a drug is therefore conservative with regards the first time a drug was tested at Phase II / III for a disease. We excluded all drugs whose earliest development year was before 2005.

From the remaining drugs, we extracted therapeutic hypotheses, as described by a combination of a drug's protein targets and the diseases the drug had been tested against. We discretized therapeutic hypotheses using the PharmaProjects assigned MeSH (\url{https://www.ncbi.nlm.nih.gov/mesh/}) and Entrez \cite{maglott2005entrez} ontology identifiers for the genes and diseases respectively. \citealt{nelson2015support} and \citealt{minikel2023refining} investigate the transition between all trial phases. We use only a subset that focuses on the \textit{in vivo} efficacy of therapeutic hypotheses. As such, we kept only the therapeutic hypotheses related to drugs tested at Phase II or III, or pre-Registration, Registration or Launched with regulatory approval. We kept only the therapeutic hypotheses where there were no drugs in active development and therefore whose clinical efficacy could be determined.

Therapeutic hypotheses that had made it to Phase II or III and have no drugs in active clinical development were assumed to have failed to demonstrate \textit{in vivo} clinical efficacy while drugs that had made it to pre-Registration and above were said to have 'succeeded'. These are the positive and negative labels in the \textit{Clinical Trial Outcomes} dataset.

In constructing the \textit{Clinical Trial Outcomes} dataset we made the assumption that ceased development is indicative of a therapeutic hypothesis failing to show efficacy in a human population. We highlight that there is likely to be noise in these negative labels: drug programmes can be prosecuted or abandoned for a range of commercial reasons rather than biological ones, drug programmes may fail because sponsors failed to identify an appropriate patient population, or drug programmes may fail for pharmacological reasons peculiar to the candidate molecule.

\section{Genetics Baseline for the Clinical Trial Outcomes Dataset}
\label{appendix-cod-genetics-baseline}

Data for the genetics baseline was downloaded from the supplementary data of \citealt{minikel2023refining} (\url{https://github.com/ericminikel/genetic_support/tree/sio/data}) and reproduced using the methodology described in \citealt{minikel2023refining}, briefly summarised below.

In the supplementary data, table \textit{assoc.tsv} contains the full set of genetic associations that were templated into natural language in Section \ref{sec:cod}. These already-curated genetic associations were filtered further as per \citealt{minikel2023refining}, removing all rows with a ``source'' of `OTG` and an ``l2g\_share'' $<$ 0.5. 

There exists ontological mismatch between sources of genetic evidence and diseases referenced in the PharmaProjects data. As such, the \textit{Clinical Trial Outcomes} dataset is joined to the genetic association data by matching exactly on gene identity, and on a measure of MeSH-MeSH similarity for diseases / traits. 

The table \textit{sim.tsv.gz} contains a full list of pairwise MeSH - MeSH similarities used in this joining of datasets. The similarity measure is a composite information criterion measure of similarity on the MeSH ontology tree; see \citealt{minikel2023refining} for details.

The continuous score for the genetics baseline for each therapeutic hypotheses in the \textit{Clinical Trial Outcomes} dataset is the maximum similarity to a genetics association across all the genetic association data, where 1 implies an exact disease-disease match and 0 means the there is no path between the entities in the MeSH ontology, or there is no genetic association data available for the gene anywhere in the genetic association data.

\section{Further Details on Creation of Evidence Annotations for Clinical Trial Outcomes Dataset}
\label{appendix-cod-evidence-annotations}

We constructed \textit{Evidence Annotations} for the \textit{Clinical Trial Outcomes} dataset by having GPT-4 annotate (as relevant or irrelevant) all evidence for 50 \textit{Clinical Trial Outcome} therapeutic hypotheses associated with trial success, as well as 50 with trial failures, both randomly sampled, resulting in 6400 query-evidence pairs (100 queries each retrieving 64 pieces of evidence) with a binary annotation. The same instructions were followed by the drug discovery expert when providing annotations used to validate the GPT-4 annotations in Section \ref{sec:cod}. The expert annotator was a Principal Scientist with over two years industry drug target identification experience in addition to holding a domain-relevant PhD and post-doc. The expert consented to the use of their annotations.

Separately and using a similar approach, we created the dataset of evidence annotations used for auditing explanations as described in Section \ref{results-auditing}. In this case, we computed \acrshort{ours} Shapley values (computed on pre-sigmoid outputs) for all retrieved evidence over all \textit{Clinical Trial Outcomes} dataset examples, ordered the evidence by Shapley value, and selected the 20,000 evidence sentences with highest Shapley values. We then ran relevant/irrelevant annotations on this subset using GPT-4.

We used the combined pre-2005 literature and templated genetics corpus for both tasks. Relevant/irrelevant annotations were obtained through the use of GPT-4, using the following prompt, substituting in DISEASE\_OF\_INTEREST and EVIDENCE\_SENTENCE:

\begin{verbatim}
    You are a scientific expert working on target identification in drug 
    discovery.
    
    Your task is to use your expertise to evaluate a piece of evidence 
    (referred to as EVIDENCE) for a potential drug target for a specified 
    disease (referred to as DISEASE). Specifically you must indicate whether 
    the EVIDENCE about a masked target (in the form of a sentence from an 
    academic paper), provides relevant evidence that the drug target might be 
    promising for developing a treatment for the DISEASE. If the EVIDENCE 
    sentence does not make any link to the biology of the specified DISEASE, 
    then it is not relevant. Please explain your reasoning first before giving 
    your answer. Provide your final answer by stating either 
    <answer>RELEVANT</answer> or <answer>IRRELEVANT</answer>. Please also 
    indicate your confidence in your answer by writing one of 
    <confidence>HIGH</confidence> or <confidence>LOW</confidence>.
    
    Here are some examples demonstrating proper formatting and reasoning in 
    a response:
    <example>
    H: Your DISEASE of interest is Sarcopenia.
    
    Here is the EVIDENCE sentence, containing a masked target, to assess:
    <evidence>
    Many studies also described exercise-induced increases in transcriptional 
    and translational levels of FGFR1, [MASK], and/or KLB [29,33,35,36].
    </evidence>
    
    A:The evidence sentence makes no indication that [MASK] plays a role in 
    Sarcopenia, therefore it is <answer>IRRELEVANT</answer> with 
    <confidence>HIGH</confidence>.
    </example>
    
    <example>
    H: Your DISEASE of interest is Amyotrophic Lateral Sclerosis.
    
    Here is the EVIDENCE sentence, containing a masked target, to assess:
    <evidence>
    Therefore, further study is needed to clarify where [MASK] functions 
    during lysosome trafficking and neurite outgrowth.
    </evidence>
    
    A:The evidence sentence implies that [MASK] may play a role in biological 
    mechanisms directly related to ALS, but the phrasing is ambiguous, therefore 
    it is <answer>RELEVANT</answer> with <confidence>LOW</confidence>.
    </example>
    
    <example>
    H: Your DISEASE of interest is Lung Adenocarcinoma.
    
    Here is the EVIDENCE sentence, containing a masked target, to assess:
    <evidence>
    Conversely, [MASK] mRNA and protein expression is higher in a variety of 
    tumor tissues, including lung cancer [ 7 , 20 ].
    </evidence>
    
    A:The evidence sentence states that [MASK] expression is upregulated in tumor 
    tissues including our specific cancer type of lung cancer, therefore it is 
    <answer>RELEVANT</answer> with <confidence>HIGH</confidence>.
    </example>
    
    
    Now here is your real task. 
    Your DISEASE of interest is {DISEASE_OF_INTEREST}.
    
    Here is the EVIDENCE sentence, containing a masked target, to assess:
    <evidence>
    {EVIDENCE_SENTENCE}
    </evidence>
\end{verbatim}

\section{Examples of Evidence Annotations for Clinical Trial Outcomes Dataset}
\label{appendix-cod-evidence-annotations-examples}

Here we show an example of each of true positive, true negative, false positive, and false negative \acrshort{ours} evidence Shapley values versus GPT-4 relevant/irrelevant \textit{Evidence Annotations}, for \textit{Clinical Trial Outcomes} data.

\begin{enumerate}
\item \textbf{True positive}: \acrshort{ours} assigns a large, positive Shapley value to a piece of evidence that GPT-4 labels as relevant. \acrshort{ours} and GPT-4 recognise this as a relevant piece of evidence.

\begin{verbatim}
    Query: [MASK] is a promising drug target for the treatment of
    myelodysplastic syndromes.

    Target: BIRC2
    
    Evidence: [MASK] protein expression was found to be present in a majority
    of leukemic blasts from untreated patients, but absent in normal 
    differentiating myeloid progenitor cells. (PMID: 14666661)

    R2E-generated Shapley value: 0.4057

    GPT4 completion: A: The evidence sentence indicates that [MASK] protein is
    expressed in leukemic blasts, which are abnormal cells associated with 
    leukemia, a type of blood cancer. Myelodysplastic Syndromes (MDS) are a 
    group of disorders caused by poorly formed blood cells or ones that don't 
    work properly, which can often progress to leukemia. The fact that [MASK] 
    is not present in normal myeloid progenitor cells, which are cells that 
    give rise to all types of blood cells, suggests that it could be involved 
    in the abnormal development of these cells in MDS. Therefore, it is 
    <answer>RELEVANT</answer> with <confidence>HIGH</confidence>.
\end{verbatim}

\item \textbf{True negative}: \acrshort{ours} assigns a negative Shapley value to a piece of evidence that GPT-4 labels as irrelevant. \acrshort{ours} and GPT-4 recognise this as an irrelevant piece of evidence.

\begin{verbatim}
    Query: [MASK] is a promising drug target for the treatment of crohn 
    disease.

    Target: LANCL2
    
    Evidence: Because of the structural similarity to LanC, we postulate that
    [MASK] may play a role as a component of a peptide-modifying complex. 
    (PMID: 11762191)

    R2E-generated Shapley value: -0.0160

    GPT4 completion: A: The evidence sentence suggests that [MASK] may be 
    involved in peptide modification, but it does not provide any direct link 
    or relevance to Crohn Disease. Therefore, it is <answer>IRRELEVANT</answer>
    with <confidence>HIGH</confidence>.
\end{verbatim}

\item \textbf{False positive}: \acrshort{ours} assigns a large, positive Shapley value to a piece of evidence that GPT-4 labels as irrelevant. \acrshort{ours} seems to misunderstand an abbreviation that GPT-4 is able to correctly parse.

\begin{verbatim}
    Query: [MASK] is a promising drug target for the treatment of chronic 
    myelomonocytic leukemia.

    Target: MAP3K1

    Evidence: This observation indicates that [MASK] may be a useful
    therapeutic target in the treatment of CML. (PMID: 14586403)

    R2E-generated Shapley value: 0.2329

    GPT4 completion: A:The evidence sentence suggests that [MASK] could
    be a potential therapeutic target for Chronic Myeloid Leukemia (CML).
    However, the disease of interest is Chronic Myelomonocytic Leukemia,
    not CML. Despite both being types of leukemia, they are distinct diseases
    with different pathologies. Therefore, the evidence is
    <answer>IRRELEVANT</answer> with <confidence>HIGH</confidence>.
\end{verbatim}

\item \textbf{False negative}: \acrshort{ours} assigns a negative Shapley value to a piece of evidence that GPT-4 labels as relevant. \acrshort{ours} fails to recognize that this is a relevant piece of evidence.

\begin{verbatim}
    Query: [MASK] is a promising drug target for the treatment of heart 
    failure.

    Target: RXFP1

    Evidence: However, a recent study has identified the relaxin receptor as a
    G-protein-coupled orphan receptor, [MASK], that can couple to adenylate
    cyclase to increase cAMP ( Hsu et al ., 2002 ). (PMID: 12381685)

    R2E-generated Shapley value: -0.0167

    GPT4 completion: A: The evidence sentence suggests that [MASK] is a
    G-protein-coupled orphan receptor that can couple to adenylate cyclase
    to increase cAMP. This is relevant to heart failure as G-protein-coupled
    receptors and cAMP signaling pathways are known to play crucial roles in
    cardiac function and heart failure pathology. Therefore, it is
    <answer>RELEVANT</answer> with <confidence>HIGH</confidence>.
\end{verbatim}
\end{enumerate}

\section{Details on Cleaning MeSH Terms for Templating}
\label{appendix-cleaning-mesh-terms}

The canonical MeSH name for each MeSH identifier is used in templates for generating \acrshort{ours} queries in the \textit{Clinical Trial Outcomes} evaluation and generating genetics-derived sentences for augmenting the literature evidence.

The MeSH names underwent light reformatting to align them to how they might be expressed in natural language. Simply, we lowercase the MeSH name, split on commas, and reverse order the resulting list. So for example the MeSH name \textit{Leukemia, Myelomonocytic, Chronic} becomes \textit{chronic myelomonocytic leukemia}.

\section{Further Results on Predicting Clinical Trial Outcomes and Genetic Evidence}
\label{appendix-cod-results}

\subsection{Relative Success}
\label{appendix-cod-rs}

For a given prediction threshold, we compute relative success of model predictions as:
\begin{equation}
RS = \frac{(\text{True Positive} / \text{Predicted Positive})}{(\text{False Negative} / \text{Predicted Negative})}
\end{equation}

Where relevant, we use Katz method \cite{katz1978} for confidence intervals and Z-test for comparisons.

\subsection{Results for Diseases with Genetic Insight}
\label{appendix-cod-genetic-insight}

Previous analyses of genetic methods for target identification have restricted to evaluating only on diseases with at least one piece of genetics data and for which therefore genetics could be expected to be informative (those with 'genetic insight') \cite{minikel2023refining}. In \citet{minikel2023refining}, diseases were deemed to have genetic insight if there was at least one genetic association between a gene and disease with a MeSH-MeSH similarity of $>$ 0.7. This subsetting of therapeutic hypotheses was used to obtain the widely published relative success of $\sim$2 in predicting clinical trial outcome success from genetic data. 

We validated our \textit{Clinical Trial Outcomes} dataset by corroborating this result by similarly restricting post-2005 therapeutic hypotheses to diseases with genetic insight, and using a MeSH-MeSH similarity threshold of $>$0.8 as the threshold for positive predictions as per \citealt{minikel2023refining}. At this threshold, the genetics baseline makes 500 positive predictions across the 4,056 therapeutic hypotheses, with a Relative Success of 1.98, 95\% CI (1.76, 2.24). In comparison, \acrshort{ours}-cor predicting on literature obtained a relative success of 2.17 (95 \% CI (2.44, 1.93)) making the same number of positive predictions.

For completeness, we also show AUROC results after restricting to diseases with genetic insight in Table \ref{cod-table-genetic-insight-hypotheses}, with trends in AUROC similar to the results without restriction shown in the main text - rationale for the latter below (Appendix \ref{appendix-cod-all-diseases}).

\begin{table*}[ht!]
\caption{\textbf{Clinical Trial Outcomes on therapeutic hypotheses with genetic insight}: AUROC for \acrshort{ours} retrieving from literature-alone, genetics-alone, or both; in comparison to baselines, when subsetting therapeutic hypotheses just to those where the disease has at least one genetic association in the genetics baseline.}
\label{cod-table-genetic-insight-hypotheses}
\begin{center}
\begin{small}
\begin{sc}
\begin{tabular}{ll|cc}
\toprule
 Model & Corpus &  AUROC \\
\midrule
Genetic         & Genetics &  0.588 \\
Freq        & Literature & 0.552 \\
MCS         & Literature & 0.634 \\
MLM         & Literature & 0.638 \\
\hline
\acrshort{ours}-uncor & Genetics &  0.618 \\
\acrshort{ours}-uncor & Literature &  0.636 \\
\acrshort{ours}-cor & Literature & 0.643 \\
\acrshort{ours}-cor & Both & \textbf{0.647} \\
\midrule
\midrule
\acrshort{ours}-audit & Both & \textbf{0.651} \\
\bottomrule
\end{tabular}
\end{sc}
\end{small}
\end{center}
\end{table*}

\subsection{Results for All Diseases}
\label{appendix-cod-all-diseases}

When comparing to predictions using literature evidence, restricting to diseases with genetic insight as described above, would undervalue literature as an evidence source; literature can be expected to be informative about a wider range of diseases. Therefore, for AUROC results in the main text (\ref{cod-table-1}) we instead show performance against all diseases in the \textit{Clinical Trial Outcomes} data, without restriction to those with genetic insight.

In Figure \ref{appendix-figure-relative-success}, we show the relative success for a given number of positive predictions for each model, by varying thresholds for each model. The relative success of the genetics baseline is below that of all \acrshort{ours} models using literature evidence, across all model thresholds (Figure \ref{appendix-figure-relative-success}), as well as largely below the \acrshort{ours} model using genetics-evidence only. As expected, compared to when restricting to diseases with genetic insight (Appendix \ref{appendix-cod-genetic-insight}), the genetics baseline (using the same $>$0.8 threshold) has a lower relative success (1.72, 95\% CI (1.54, 1.93)) when predicting for all diseases.

\subsection{GPT-4-FS-RAG-CoT Baseline}
\label{appendix-cod-gpt-baseline-results}

\begin{figure}[ht!]
\centering
\includegraphics[width=0.7\linewidth]{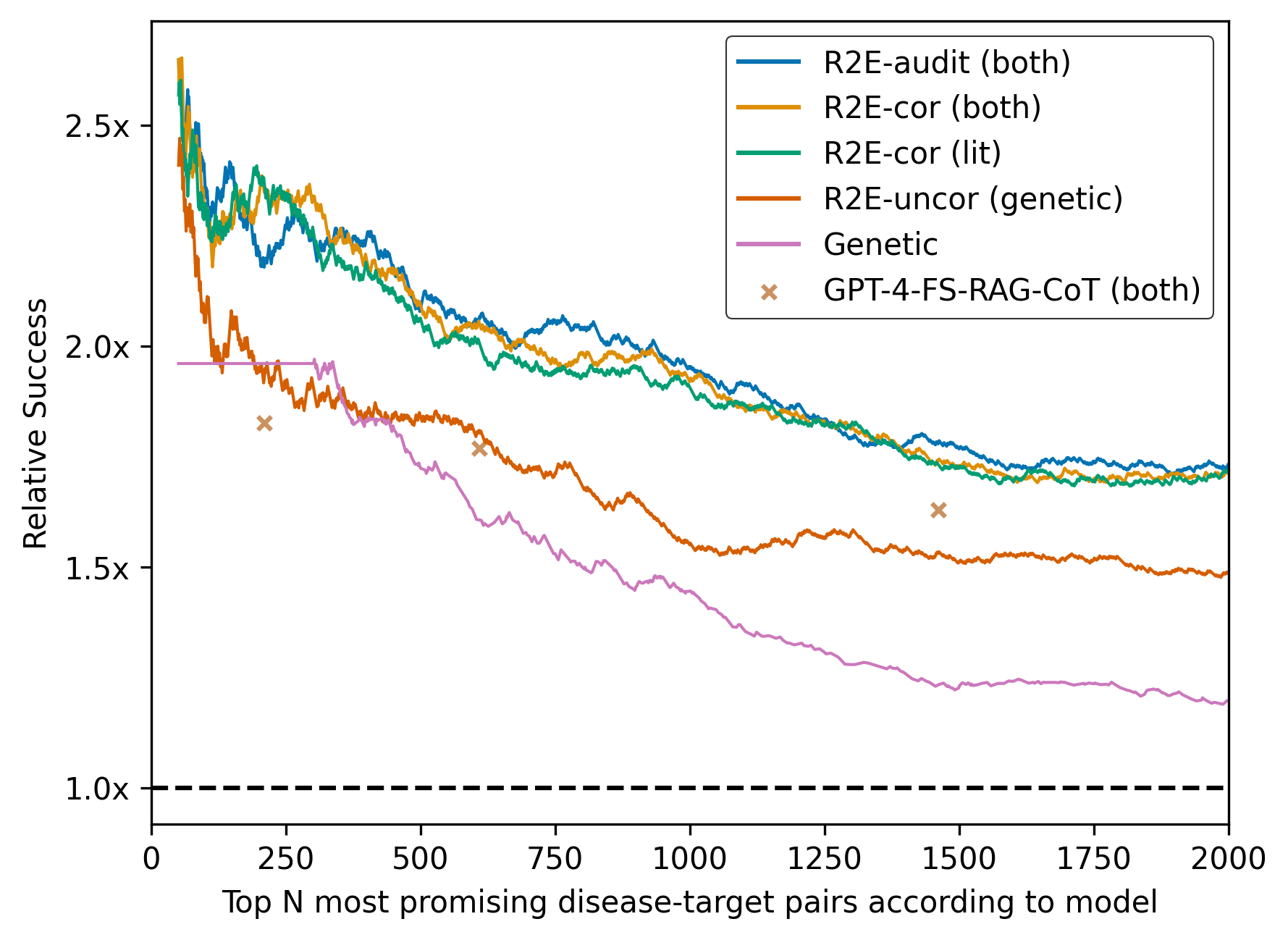}
\caption{\textbf{\acrshort{ours} Relative Success on Clinical Trial Outcomes}. Relative success for a given number of positive predictions (x-axis) for each model. The different numbers of positive predictions was achieved by varying the threshold for a positive prediction for each model.}
\label{appendix-figure-relative-success}
\end{figure}

The few-shot, chain-of-thought prompted GPT-4 baseline with retrieval augmentation (see Appendix \ref{appendix-gpt-baseline} for details of setup) had lower relative success than all \acrshort{ours} models using literature evidence, at all thresholds (Figure \ref{appendix-figure-relative-success}). When matching thresholds to obtain 609 positive predictions, \acrshort{ours}-cor (both) (relative success: 2.05; 95\% CI (1.86, 2.26)) significantly outperformed the GPT-4 baseline (relative success: 1.77; 95\% CI (1.59, 1.97)) using the same evidence (Z test, $p=0.043$).

\subsection{Performance by Disease Area}
\label{appendix-cod-by-disease-area}

\begin{figure}[ht!]
\centering
\includegraphics[width=0.65\linewidth]{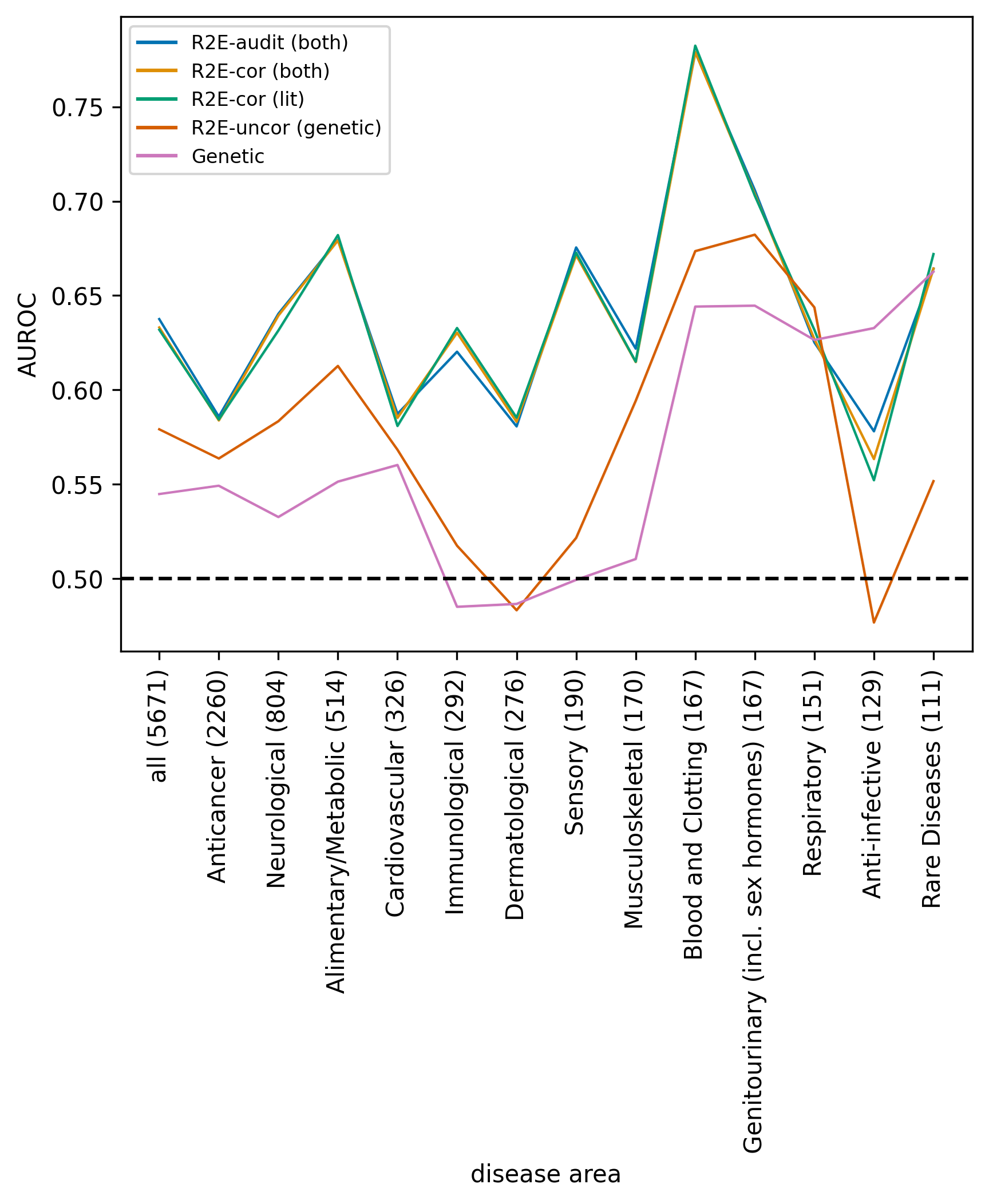}
\caption{\textbf{\acrshort{ours} performance across disease areas}. AUROC in each PharmaProjects annotated disease area with more than 100 therapeutic hypotheses. Predictions by \acrshort{ours} retrieving from literature-alone (\acrshort{ours}-cor (lit)), genetics-alone (\acrshort{ours}-uncor (genetic)), both genetics and literature (\acrshort{ours}-cor (both)), or genetics and literature with LLM auditing (\acrshort{ours}-audit (both)); in comparison to the genetics baseline (Genetic). The number of therapeutic hypotheses for each disease area are given in brackets.}
\label{appendix-figure-disease-area-auc}
\end{figure}

Figure \ref{appendix-figure-disease-area-auc} shows that there is substantial variation in performance across disease areas and modality. The variability is especially pronounced for the genetics baseline and \acrshort{ours} using only genetics-evidence, consistent with the reduced disease coverage of genetics compared to the literature. The magnitude of difference in performance between \acrshort{ours} retrieving from genetics alone and \acrshort{ours} retrieving from literature, varies by disease area. This may indicate disease areas for which alternative predictive modalities to genetics might be being represented in the literature.

\subsection{Results with Forced \acrshort{ours} Retrieval of Genetics}
\label{appendix-cod-combined-retrieval-approaches}

Table \ref{cod-table-multimodality-method} shows that in the multi-modal context (with a corpus of sentences from the biomedical literature and from the genetics data), forcing retrieval of genetics evidence does not change the AUROC.

\begin{table*}[ht!]
\caption{\textbf{Methods of multimodality for Clinical Trial Outcomes}: AUROC for \acrshort{ours}-uncor and \acrshort{ours}-cor with three different methods of multi-modalility: (1) Retrieve from a single corpus containing both genetics and literature sentences (single index); (2) Retrieve up to four sentences from the genetics corpus - where possible - and retrieve the remaining sentences from the literature corpus (separate index); and (3) \acrshort{ours} scores evidence from the genetics and the literature corpora separately and the final score is the mean of the two (post-hoc aggregation).}
\label{cod-table-multimodality-method}
\vskip 0.15in
\begin{center}
\begin{small}
\begin{sc}
\begin{tabular}{lll|cc}
\toprule
Model & Corpus & Method & AUROC \\
\hline
\acrshort{ours}-uncor & Both & Single index & 0.631 \\
\acrshort{ours}-cor & Both & Single index & 0.633 \\
\acrshort{ours}-uncor & Both & Separate index & 0.631 \\
\acrshort{ours}-cor & Both & Separate index & 0.633 \\
\acrshort{ours} & Both & Post-hoc aggregation & 0.633 \\
\bottomrule
\end{tabular}
\end{sc}
\end{small}
\end{center}
\vskip -0.1in
\end{table*}

\newpage

\subsection{\acrshort{ours} Benefits from Soft Semantic Matching}
\label{appendix-cod-semantic-matching}

In evidence auditing experiments detailed in Section \ref{results-auditing}, where high Shapley value evidence sentences were annotated by GPT-4 as relevant or irrelevant to the given query, 527/809 of the annotated genetics sentences were annotated as relevant, evidencing that \acrshort{ours} can appropriately leverage genetic evidence. Note that 268 of these 527 genetic evidence sentences was related by \acrshort{ours} to a disease that was neither a substring of, nor contained, the \textit{Clinical Trial Outcomes} disease.

\begin{figure}[ht!]
\centering
\includegraphics[width=0.7\linewidth]{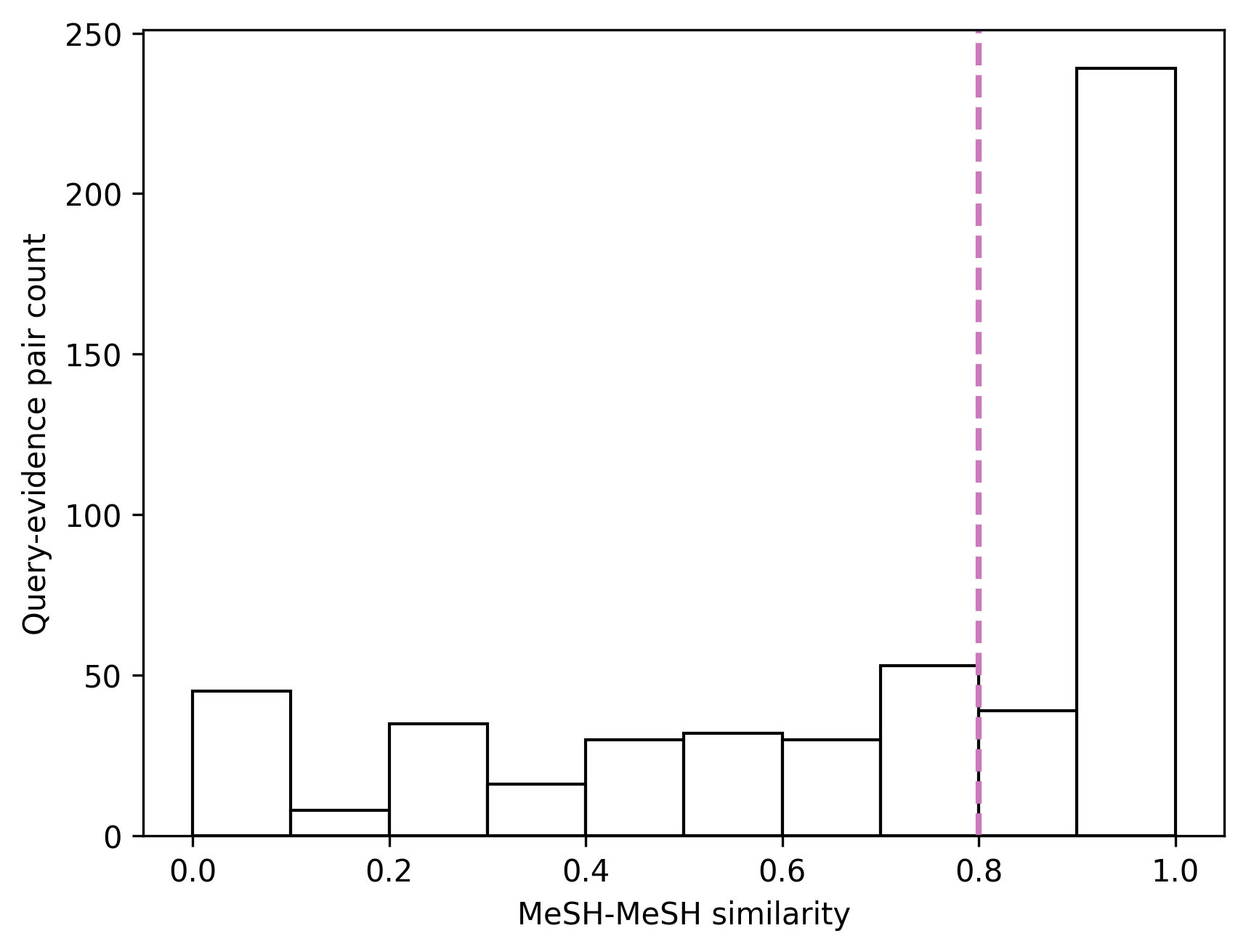}
\caption{\textbf{MeSH-MeSH ontological similarity of query-genetics evidence pairs with high Shapley scores}. Histogram showing the distribution of similarities between MeSH terms for diseases / traits in genetics evidence annotated as ``Relevant'' by GPT-4, and diseases in the clinical trial therapeutic hypothesis queries, when the evidence had a high Shapley score (Section \ref{results-auditing}). The dashed line indicates the threshold used in the \citealt{minikel2023refining} genetics baseline to assign genetic evidence to a therapeutic hypothesis.}
\label{appendix-figure-trait-trait-similarity}
\end{figure}

Figure \ref{appendix-figure-trait-trait-similarity} shows the distribution of MeSH-MeSH ontological similarity, between the clinical trial disease and the genetics evidence disease / trait, as calculated by \citealt{minikel2023refining}, for these 527 relevant-annotated genetic query-evidence pairs with high Shapley scores (Section \ref{results-auditing}). Note that when calculating relative success in \citealt{minikel2023refining}, the threshold MeSH-MeSH similarity for positively linking between therapeutic hypotheses and genetic association data was 0.8. By contrast, we observed that \acrshort{ours} can also perform ``soft'' semantic matching between the query and the genetics evidence. For example, \acrshort{ours} picked up on the following trait-trait pairs with a MeSH similarity $<$ 0.2: (erythrocyte count and anemia), (eosinophilia and asthma), (astrocytoma and brain neoplasms). This highlights the shortcomings of a universal threshold based on ontological similarity metrics (genetics baseline) versus semantic matching and reasoning through natural language (\acrshort{ours}).

\section{A Few-Shot, Chain-of-Thought, RAG Baseline using GPT-4 on Clinical Trial Outcomes}
\label{appendix-gpt-baseline}

\subsection{Existing LLMs and Retrieval Augmented Generation}
\label{appendix-existing-llms}

We first detail why in general, to the best of our knowledge, generative LLMs such as GPT-4, do not solve the problem we seek to address with \acrshort{ours}, of explainable multi-label prediction from textual data, i.e.:
\begin{itemize}
    \item Score and rank each answer in the answer set
    \item Faithfully attribute the score to evidence passages
\end{itemize}

With access to token probabilities, one option could be to constrain LLM completions to synonyms of entities and compute completion probabilities to rank entities. For explainability one might then consider retrieval augmented generation (RAG). However, question-answering-style generative explanations are often not faithful and are not quantitative - they do not assess the direct, quantitative impact of a piece of evidence on the model score, and they risk hallucination. Additionally, running LLMs with separate gene-specific prompts in a RAG setup, for each of 19,176 genes for every query, would be prohibitively expensive.

For models restricted to API access only, such as GPT-4, it is not possible to use completion probabilities to rank genes. Prompting to directly generate a ranked list of targets returns well-known targets and omits explainability. Using a RAG approach for each gene independently, one could have the model specify a score to partially rank targets, or at least identify a subset of targets the LLM deems promising. However, once again this does not ensure faithful explainability and would be similarly prohibitively expensive for ranking 19,176 genes for every query.

\subsection{Setup for Comparison to GPT-4 Baseline}

Despite the points raised above on prohibitive expense (Appendix \ref{appendix-existing-llms}), in the case of \textit{Clinical Trial Outcomes}, since the evaluation only requires prediction on individual disease-target pairs, rather than full rankings of all genes for each query, a direct comparison to the latter RAG-based approach using GPT-4 is possible. Despite the described lack of faithful explainability and the practical inappropriateness of this approach to the problem addressed by \acrshort{ours}, for academic interest only we include a comparison. We also highlight that since it is not possible to use a GPT-4 model only trained on pre-2005 data, it may be advantaged in comparison to \acrshort{ours} when evaluating on our dataset of \textit{Clinical Trial Outcomes} from 2005 onwards.

Specifically, we use a chain-of-thought, few-shot prompted GPT-4 in a RAG style setup. For each disease-target pair we provide the same evidence set of up to 64 sentences as seen by \acrshort{ours}, and have GPT-4 predict whether the evidence supports the masked target as promising or not for developing a treatment for the given disease, as well as a level of confidence in the prediction out of very low, low, medium, high, very high. We summarise the findings in Appendix \ref{appendix-cod-gpt-baseline-results}, and show the results in terms of relative success in Figure \ref{appendix-figure-relative-success} with the following 3 different cutoffs used to determine positive predictions, corresponding to the three data points shown in the figure:
\begin{itemize}
    \item At least very low confidence in the target being promising
    \item At least high confidence in the target being promising
    \item At least very high confidence in the target being promising
\end{itemize}

\subsection{Prompting for GPT-4 Baseline}

The following few-shot, chain-of-thought, retrieval-augmented prompt was used for the GPT-4 baseline, where we substitute {DISEASE OF INTEREST} and {EVIDENCE SENTENCES} for the particular evaluation query. The PMIDs included inline in this prompt are not passed to GPT-4, but are included in order to properly reference these works in this manuscript.

\begin{verbatim}
    You are a scientific expert working on target identification in drug 
    discovery.
    
    Your task is to use your expertise to evaluate whether a potential drug 
    target could potentially be promising for a given disease (referred to as 
    DISEASE). You must make your evaluation based on a provided set of evidence 
    about the drug target (referred to as EVIDENCE), identifying if any of the 
    EVIDENCE could directly or indirectly suggest the target could be promising.
    
    Please explain your reasoning first before giving your answer. 
    
    Provide your final answer by stating either <answer>PROMISING</answer> or 
    <answer>NOT PROMISING</answer>.
    
    Please also indicate your confidence in your answer by writing one of:
    - <confidence>VERY HIGH</confidence>
    - <confidence>HIGH</confidence>
    - <confidence>MEDIUM</confidence>
    - <confidence>LOW</confidence>
    - <confidence>VERY LOW</confidence>.
    
    Note that the name of the target will be hidden in the EVIDENCE set. Mentions 
    of the target have been replaced with '[MASK]'. This is because you should 
    make your prediction based on the evidence itself, not based on the particular 
    target.
    
    Here are some illustrative examples of the task demonstrating proper 
    formatting and reasoning in a response.
    
    <example>
    
    TASK: Your DISEASE of interest is lung adenocarcinoma.
    
    Here is the set of EVIDENCE about the target:
    <evidence>
    1. Thus, [MASK] is also a novel prognostic biomarker and therapeutic target 
    for NSCLC. [PMID: 36215859]
    2. In the present study, we demonstrated that [MASK] was significantly 
    upregulated in tumor tissues and associated with poor clinical prognosis of 
    NSCLC. [PMID: 32855383]
    3. As expected, EMT-related gene sets were significantly enriched in the 
    [MASK]-high expression phenotype, suggesting that [MASK] may contribute to 
    TGF-β-induced EMT of NSCLC cells (Supplementary Fig. S6A). [PMID: 32855383]
    4. [MASK] is highly expressed in non small cell lung cancer tissues and is 
    associated with poor prognosis. [PMID: 32167655]
    5. Collectively, this study supports that [MASK] is a key regulator in 
    IL-6/JAK2/STAT3 axis and mediates EGFR inhibitor resistance in lung 
    adenocarcinoma. [PMID: 36990047]
    6. The cell growth was suppressed after [MASK] was knocked out in 
    established PC9 sg-[MASK] cells, which confirmed that [MASK] is essential 
    for cell survival of NSCLC (Figure S1I). [PMID: 31607564]
    </evidence>
    
    Please evaluate whether any of the provided EVIDENCE suggests that the target 
    could be promising for lung adenocarcinoma.
    
    RESPONSE: [MASK] is increased in NSCLC (a subtype of lung adenocarcinoma) 
    tissues and is associated with poor survival (EVIDENCE: 2, 4). [MASK] is linked 
    to relevant NSCLC mechanisms including EMT and EGFR resistance (EVIDENCE: 3, 6). 
    There is also supporting assay data to suggest inhibiting [MASK] would prevent 
    NSCLC cell growth (EVIDENCE: 1).
    
    Conclusion: <answer>PROMISING</answer>.
    Confidence: <confidence>VERY HIGH</confidence>.
    </example>
    
    <example>
    
    TASK: Your DISEASE of interest is multiple sclerosis.
    
    Here is the set of EVIDENCE about the target:
    <evidence>
    1. Given, that [MASK] does not have functionality in the MHC-II antigen 
    presenting pathway, it is possible that [MASK] promotes MS pathogenesis via 
    inflammasome activation. [PMID: 30817945]
    2. [MASK] is an emerging pharmacological target for cancer immunotherapy and 
    the control of inflammatory autoimmunity, including rheumatic conditions such 
    as AS (36, 37). [PMID: 33617882]
    3. A Functional Variant in [MASK] Predisposes to Multiple Sclerosis 
    [PMID: 22253828]
    4. In the light of foregoing discussion [MASK] can be envisaged as a relevant 
    target for prevention and treatment of autoimmune diseases. [PMID: 36740089]
    5. In this study, we present, to our knowledge, the first mechanistic studies 
    performed to uncover why polymorphisms in [MASK] are associated with increased 
    susceptibility to MS. [PMID: 34810226]
    </evidence>
    Please evaluate whether any of the provided EVIDENCE suggests that the target 
    could be promising for multiple sclerosis.
    
    RESPONSE: There is a possible mechanistic link from [MASK] to multiple 
    sclerosis pathogenesis via inflammasome activation (EVIDENCE: 1). [MASK] 
    is genetically linked to multiple sclerosis (EVIDENCE: 3), which is 
    potentially via a mechanistic function (EVIDENCE: 5). [MASK] has been 
    described as a therapeutic target for similar autoimmune diseases 
    (EVIDENCE: 2, 4).
    
    Conclusion: <answer>PROMISING</answer>.
    Confidence: <confidence>HIGH</confidence>.
    </example>
    
    <example>
    
    TASK: Your DISEASE of interest is idiopathic pulmonary fibrosis.
    
    Here is the set of EVIDENCE about the target:
    <evidence>
    1. The antimicrobial peptide YD attenuates inflammation via miR-155 targeting 
    [MASK] during liver fibrosis. [PMID: 33532183]
    2. Although [MASK]−/− mice reacted similarly to WT mice when allowed to 
    recover from an acute DSS-induced injury ( Figure 1) and exhibited signs of 
    improved repair ( Figure 2), they had an increased inflammatory response 
    compared to WT animals ( Figures 5A and 5B ). [PMID: 20226691]
    3. Consistent with their response to acute DSS treatment and their enhanced 
    tissue repair phenotype, [MASK]−/− mice were more resistant to chronic colitis 
    compared to WT animals, gaining weight by the end of the experiment as compared 
    to WT mice that lost 5% of their initial body weight ( Figure 5C). 
    [PMID: 20226691]
    4. [MASK]−/− mice showed a comparable phenotype to WT mice in the acute model 
    of DSS colitis, but expressed an increased mortality when DSS exposure was 
    prolonged to 15 days. [PMID: 20346770]
    5. Altogether, these data suggested that [MASK]−/− mice have an increased 
    ability to recruit macrophages, which leads to increased production of 
    inflammatory and tissue repair factors. [PMID: 20226691]
    6. Yan et al. [ 301 ] recently reported that the anti-fibrotic properties of 
    AMP YD were mediated through the miR-155/[MASK]/NF-kB pathway. 
    [PMID: 34496967]
    7. [MASK] is an inhibitor of caspase 1, and Dupaul-Chicoine et al . showed 
    that [MASK] −/− mice are resistant to acute and chronic (but not sustained) 
    DSS-induced colitis [PMID: 20425920]
    </evidence>
    
    Please evaluate whether any of the provided EVIDENCE suggests that the target 
    could be promising for idiopathic pulmonary fibrosis.
    
    RESPONSE: The evidence largely points to [MASK] having a role in inflammation 
    rather than specifically fibrosis (EVIDENCE: 2, 3, 4, 5, 7). None of the 
    evidence specifies that [MASK] is expressed in the lung and none of the 
    evidence provides direct support for the role of [MASK] in IPF. However, there 
    is indirect linking of [MASK] to fibrosis via an indirect mechanism in a 
    different disease context to idiopathic pulmonary fibrosis (EVIDENCE: 1, 6).
    
    Conclusion: <answer>NOT PROMISING</answer>.
    Confidence: <confidence>LOW</confidence>.
    </example>
    
    Now here is your real task.
    
    Your DISEASE of interest is {DISEASE OF INTEREST}.
    
    Here is the set of EVIDENCE about the target:
    <evidence>
    {EVIDENCE SENTENCES}
    </evidence>
    
    Please evaluate whether any of the provided EVIDENCE suggests that the target 
    could be promising for {DISEASE OF INTEREST}.  
\end{verbatim}

\section{Examples of Auditing of Evidence for Clinical Trial Outcomes Dataset}
\label{appendix-cod-auditing-examples}

Here we show examples of query-evidence pairs that GPT-4 annotated as irrelevant and to which \acrshort{ours} assigned a large positive Shapley value, as identified during the auditing experiments detailed in Section \ref{results-auditing}. With each example we also report the overall \acrshort{ours} prediction score for the associated query-target pair before and after the auditing process. The change in score from before to after the auditing process accounts for the removal of the specified ``false positive'' (according to GPT-4) piece of evidence, as well as any other query-evidence ``false positive'' pairs removed by the audit. We also include the final clinical trial status of the associated query-target pair in each example.

\begin{enumerate}
\item \textbf{Negative evidence example}: GPT-4 correctly identifies the negation in the evidence sentence, which suggests that there is not good evidence for the involvement of [MASK] in Tourette's syndrome. \acrshort{ours} fails in this case and assigns a large positive Shapley value to this irrelevant piece of evidence.

\begin{verbatim}
    Query: [MASK] is a promising drug target for the treatment of tourette 
    syndrome.

    Target: HTR1A

    Clinical Trial Outcome: Failure

    R2E score pre-audit: 0.9831

    R2E score post-audit: 0.9395
    
    Evidence: In conclusion, our study does not provide evidence that the 
    [MASK] gene plays either a major or a minor role in the genetic 
    predisposition to schizophrenia, bipolar affective disorder, or Tourette's 
    syndrome. (PMID: 8546152)

    R2E-generated Shapley value: 0.0371

    GPT4 completion: A:The evidence sentence explicitly states that the [MASK] 
    gene does not play a major or minor role in the genetic predisposition to 
    Tourette's syndrome, therefore it is <answer>IRRELEVANT</answer> with 
    <confidence>HIGH</confidence>.
\end{verbatim}

\item \textbf{Failed entity linking example}: \acrshort{ours} incorrectly understands "ICH" to refer to the disease of interest "ichytosis", which is a skin disorder, and therefore assigns a large Shapley value to this irrelevant piece of evidence. GPT-4 correctly infers that "ICH" refers here to intracerebral hemorrhage based on the surrounding context in the evidence sentence, and is therefore irrelevant to ichytosis.

\begin{verbatim}
    Query: [MASK] is a promising drug target for the treatment of ichthyosis.

    Target: TNF

    Clinical Trial Outcome: Success

    R2E score pre-audit: 0.8920

    R2E score post-audit: 0.7040
    
    Evidence: [MASK] likely plays a central and pathogenic role during ICH,
    because antisense oligonucleotides that target [MASK] reduced [MASK] 
    protein levels and neuronal cell death within the hematoma as well as 
    neurobehavioural deficits following ICH. (PMID: 11409424)

    R2E-generated Shapley value: 0.1088

    GPT4 completion: A: The evidence sentence suggests that [MASK] plays a 
    central and pathogenic role during ICH, and that reducing [MASK] protein 
    levels can alleviate symptoms. However, the ICH referred to in the 
    sentence is likely intracerebral hemorrhage, not ichthyosis. Therefore, 
    the evidence is <answer>IRRELEVANT</answer> with 
    <confidence>HIGH</confidence>.
\end{verbatim}

\item \textbf{Incorrect understanding example}: \acrshort{ours} assigns a large positive Shapley value to a piece of evidence that seems to have no relevance (perhaps reasoning incorrectly based on the shared "retin" token in the disease name and the compound name). GPT-4 correctly identifies the evidence as irrelevant.

\begin{verbatim}
    Query: [MASK] is a promising drug target for the treatment of retinitis 
    pigmentosa.

    Target: NGFR

    Clinical Trial Outcome: Failure

    R2E score pre-audit: 0.6419

    R2E score post-audit: 0.3103
    
    Evidence: [MASK] was solubilized in methylene chloride and fluorinated 
    using diethylaminosulfurtrifluoride (DAST) as the fluorinating agent 
    (Fig. 1) asdescribed for the synthesis of all- trans-retinoyl-fluoride 
    [25]. (PMID: 11762191)

    R2E-generated Shapley value: 0.3807

    GPT4 completion: A: The evidence sentence describes a chemical process 
    involving [MASK], but does not link this process to the biology or 
    pathology of Retinitis Pigmentosa. Therefore, it is 
    <answer>IRRELEVANT</answer> with <confidence>HIGH</confidence>.
\end{verbatim}

\item \textbf{Indirect evidence example}: GPT-4 is arguably too restrictive in its definition of relevance here. B-cell chronic lymphocytic leukemia (B-CLL) and mantle-cell lymphoma share enough significant disease biology that this is still a potentially relevant piece of evidence. This bias against indirect evidence was an issue we saw in several auditing examples we examined, and could be mitigated by adjusting the prompting strategy to GPT-4 in future work.  

\begin{verbatim}
    Query: [MASK] is a promising drug target for the treatment of mantle-cell 
    lymphoma.

    Target: TNFRSF10B

    Clinical Trial Outcome: Failure

    R2E score pre-audit: 0.5400

    R2E score post-audit: 0.0381

    Evidence: Cells from 10 B-CLL patients previously studied by flow 
    cytometry and from an additional three B-CLL patients (L-244, L-243 and 
    L-249) were also positive for [MASK] protein by Western blot analysis 
    ( Figure 3 ). (PMID: 11753607)

    R2E-generated Shapley value: 0.5914

    GPT4 completion: A: The evidence sentence indicates that [MASK] protein is 
    present in B-CLL (B-cell Chronic Lymphocytic Leukemia) patients. However, 
    B-CLL is a different type of lymphoma than Mantle-Cell Lymphoma. While they 
    are both types of B-cell lymphomas, they have different characteristics and 
    may not share the same drug targets. Therefore, the evidence does not 
    directly link [MASK] to Mantle-Cell Lymphoma. It is 
    <answer>IRRELEVANT</answer> with <confidence>HIGH</confidence>.
\end{verbatim}
\end{enumerate}

\end{document}